\documentclass[runningheads]{llncs}
\usepackage{graphicx}
\usepackage{comment}
\usepackage{amsmath,amssymb} % define this before the line numbering.
\usepackage{color}
\usepackage[caption=false]{subfig}
\usepackage{textcomp}

\usepackage{hyperref}
\usepackage{epsfig}
\usepackage{algorithm}
\usepackage{setspace}
\usepackage{algpseudocode}
\usepackage{dsfont}
\usepackage{etoolbox}\AtBeginEnvironment{algorithmic}{\scriptsize} 

\usepackage{booktabs}
\usepackage{comment}
\usepackage{float}
\usepackage{adjustbox}
\usepackage{array}
\usepackage{ulem}
\usepackage{xcolor}

\newcommand{\spc}[2][c]{%
  \begin{tabular}[#1]{@{}c@{}}#2\end{tabular}}

\makeatletter
\newlength\mylena
\newlength\mylenb
\makeatother
\begin{document}
% \linenumbers
\pagestyle{headings}
\mainmatter
\def\ECCVSubNumber{4396}  % Insert your submission number here

\title{MPCC: Matching Priors and Conditionals for Clustering} % Replace with your title

\begin{comment}
\titlerunning{ECCV-20 submission ID \ECCVSubNumber} 
\authorrunning{ECCV-20 submission ID \ECCVSubNumber} 
\author{Anonymous ECCV submission}
\institute{Paper ID \ECCVSubNumber}
\end{comment}
%******************

%\begin{comment}
\titlerunning{MPCC: Matching Priors And Conditional for Clustering}
\author{Nicolás Astorga \inst{1, 4} \orcidID{0000-0001-6172-9362} \and
Pablo Huijse \inst{2, 4} \orcidID{0000-0003-3541-1697} \and
Pavlos Protopapas \inst{3}
%\orcidID{2222--3333-4444-5555}}
\and
Pablo Estévez \inst{1, 4} \orcidID{0000-0001-9164-4722}}

\authorrunning{N. Astorga, P. Huijse, P. Protopapas, P. Estévez.}
\institute{Department of Electrical Engineering, Universidad de Chile \and 
Informatics Institute, Faculty of Engineering Sciences, Universidad Austral de Chile \and
Institute for Applied Computational Science, Harvard University \and 
Millennium Institute of Astrophysics, Chile\\
\email{nicolas.astorga.r@ug.uchile.cl, phuijse@inf.uach.cl,  pavlos@seas.harvard.edu, pestevez@ing.uchile.cl}}
%\end{comment}
%******************
\maketitle

\begin{abstract}
% In ECCV the abstract should be 150 words long
Clustering is a fundamental task in unsupervised learning that depends heavily on the data representation that is used. Deep generative models have appeared as a promising tool to learn informative low-dimensional data representations. We propose Matching Priors and Conditionals for Clustering (MPCC), a GAN-based model with an encoder to infer latent variables and cluster categories from data, and a flexible decoder to generate samples from a conditional latent space. With MPCC we demonstrate that a deep generative model can be competitive/superior against discriminative methods in clustering tasks surpassing the state of the art over a diverse set of benchmark datasets. Our experiments show that adding a learnable prior and augmenting the number of encoder updates improve the quality of the generated samples, obtaining an inception score of $9.49 \pm 0.15$ and improving the Fr\'echet inception distance over the state of the art by a $46.9\%$ in CIFAR10.

%Clustering is a fundamental task in unsupervised learning that depends heavily on the data representation that is used. Deep generative models have appeared as a promising tool to learn informative low-dimensional data representations. Although there is an extensive literature for representation learning using deep generative models, the connection with clustering has not been explored much. We propose Matching Priors and Conditionals for Clustering (MPCC), a GAN-based model with an encoder to infer latent variables and cluster categories from data, and a flexible decoder to generate samples from a conditional latent space.We evaluate MPCC over a diverse set of benchmark datasets surpassing the state of the art in terms of clustering accuracy and in the quality of the generated samples. \ph{Give percentages}
\end{abstract}

\section{Introduction}

Clustering is a fundamental unsupervised learning problem that aims to group the input data  based on a similarity criterion. Traditionally, clustering models are trained on a transformed low-dimensional version of the original data obtained via feature-engineering or dimensionality reduction \textit{e.g.} PCA. Hence the performance of clustering relies heavily on the quality of the feature space representation. In recent years deep generative models have been successful in learning low-dimensional representations from complex data distributions, and two particular models have gained wide attention: The Variational Autoencoder (VAE) \cite{kingma2013vae}, \cite{rezende2014vae2} and the Generative Adversarial Network (GAN) \cite{goodfellow2014gan}.

In VAE an encoder and a decoder network pair is trained to map the data to a low-dimensional latent space, and to reconstruct it back from the latent space, respectively. The encoder is used for inference while the decoder is used for generation. The main limitations  of the standard VAE are the restrictive assumptions associated with the explicit distributions of the encoder and decoder outputs. For the latter this translates empirically as loss of detail in %\pe{se suponen imagenes?} \ph{It isn't only for images, what do you think about "loss of detail" instead of blurring to make it sound more general?} of
the generator output. In GAN a generator network that samples from latent space is trained to mimic the underlying data distribution while a discriminator is trained to detect whether the generated samples are true or synthetic (fake). This adversarial training strategy avoids explicit assumptions on the distribution of the generator, allowing GANs to produce the most realistic synthetic outputs up to date \cite{brock2018large}, \cite{karras2018progressive}, \cite{styleGAN}. The weaknesses of the standard GAN are the lack of inference capabilities and the difficulties associated with training (\textit{e.g.} mode collapse).

One would like to combine the strengths of these two models, \textit{i.e.} to be able to infer the latent variables directly from data and to have a flexible decoder that learns faithful data distributions. Additionally, we would like to train simultaneously for feature extraction and clustering as this performs better according to \cite{DEC}, \cite{yang2016jule}, \cite{yang2017towards}. 
Extensions of the standard VAE that modify the prior distribution to make it suitable for clustering have been proposed in \cite{SVAE}, \cite{dilokthanakul2016deep},  \cite{jiang2017vade}, although they still suffer from too restrictive generator models. On the other hand, the standard GAN has been extended to infer categories \cite{springenberg2015catgan}, \cite{tripleGAN}, \cite{badGAN}%%% with a focus on the semi-supervised learning setting 
. Other works have extended GAN to infer the posterior distribution of the latent variables reporting good results in both reconstruction and generation \cite{chen2016infogan}, \cite{dumoulin2016ali}, \cite{donahue2016bigan}, \cite{srivastava2017veegan}, \cite{li2017alice}, \cite{li2019aim}. These models do inference and have flexible generators but were not designed for clustering.%%% and may not learn representations appropriate for this task.

In this paper we propose a model able to learn good representations for clustering in latent space. The model is called Matching Priors and Conditionals for Clustering (MPCC). This is a GAN-based  model with (a) a learnable mixture of distributions as prior for the generator, (b) an encoder to infer the latent variables from the data and (c) a clustering network to infer the cluster membership from the latent variables. Code are available at \href{http://github.com/jumpynitro/MPCC}{\texttt{github.com/jumpynitro/MPCC}}.

\section{Background}

\label{background}

MPCC is based on a matching joint distribution optimization framework. %In this section we will describe the related models to understand their differences better and the motivation of our work. 
Let us denote $q(x)$ as the true distribution and $p(z)$ the prior, where $x \in \mathcal{X}$ is the observed variable and $z \in \mathcal{Z}$ is the latent variable, respectively. %$q(x)$ stands for the marginalization of the inference model $q(x,z)$ and $p(z)$ is the marginalization of the generative model $p(x,z)$ 
$q(x)$ and $p(z)$ stand for the marginalization of the inference model $q(x,z)$ and generative model $p(x,z)$, respectively%\ph{name x the observed variable and z the latent variable}
. If the joint distributions $q(x,z)$ and $p(x,z)$ match then it is guaranteed that all the conditionals  and marginals also match. Intuitively this means that we can reach one domain starting from the other, \textit{i.e.}, we have an encoder that allows us to reach the latent variables $p(z) \approx q(z) = \mathbb{E}_{q(x)}[q(z|x)]$ 
and  a generator that approximates the real distribution $q(x) \approx p(x) = \mathbb{E}_{p(z)}[p(x|z)]$. Notice that the latter approximation corresponds to a GAN  optimization problem. In the case of vanilla GAN the Jensen-Shannon divergence $D_{JS}(q(x)||p(x))$ is minimized, but other distances can be used \cite{WGAN}, \cite{IWAN}, \cite{fGAN}, \cite{LeastGAN}.

%((Although other classifications can be done \cite{lagrange} to the best of our knowledge we can classify 
Although other classifications can be done \cite{lagrange}, we recognize that the joint distribution matching problem can be divided in three general categories:
%Although other classifications can be done \cite{lagrange} we propose to divide the joint distribution matching optimization methods in three general categories:
i) matching the joints directly, ii) matching conditionals in $\mathcal{Z}$ and marginals in $\mathcal{X}$, and iii) matching conditionals in $\mathcal{X}$ and marginals in $\mathcal{Z}$. The straight forward approach is to minimize the distance between the joint distributions using a fully adversarial optimization such as \cite{dumoulin2016ali}, \cite{donahue2016bigan}, \cite{bigbigan}, which yields competitive results but still shows difficulties in reconstruction tasks likely affecting unsupervised representation learning. According to \cite{li2017alice} these issues are related to the lack of an explicit optimization of the conditional distributions.
Recent works \cite{distributionMatching}, \cite{li2019aim}, \cite{lagrange} have shown that the VAE \cite{kingma2013vae} loss function (ELBO) is related to matching the inference and generative joint distributions. This can be demonstrated for the Kullback–Leibler (KL) divergence of $p$ from $q$, which we refer as forward KL, as follows:
% In particular VAE loss optimization function can be obtained as a Kullback–Leibler divergence minimization of the posteriors and marginals distribution as observed in equation \ref{vae_match}.
\begin{align}
    & D_{KL}(q(z, x) ||p(z, x)) \nonumber \\ 
    ~ = ~ &\mathbb{E}_{q(x)} [D_{KL}(q(z|x) ||p(z|x))] + D_{KL}(q(x) ||p(x)) \nonumber \\
    ~ = ~ &\mathbb{E}_{q(x)} \mathbb{E}_{q(z|x)}[  -  \log p(x|z)    ]  +  \mathbb{E}_{q(x)} [D_{KL}(q(z|x)|| p(z)) ]  + \mathbb{E}_{q(x)} [\log q(x)] \nonumber \\ 
    ~ = ~ &\mathbb{E}_{q(x)} [-\text{ELBO}]  + \mathbb{E}_{q(x)} [\log q(x)] , 
    \label{vae_match}
\end{align}
hence maximizing the ELBO can be seen as matching the conditionals in latent space $\mathcal{Z}$ and the marginals in data space $\mathcal{X}$ (see the second line in Eq. \ref{vae_match}). The proof for the first equivalence in Eq. \ref{vae_match} can be found in the Appendix A.

In order to avoid latent collapse and the parametric assumptions of VAE, AIM \cite{li2019aim} proposed the opposite, \textit{i.e.} to match the conditionals in data space and the marginals in latent space. Starting from the KL divergence of $q$ from $p$, which we refer as reverse KL, %between the joint distributions 
they obtained the following:
%Notice that in equation \ref{vae_match} the distance between the marginal distributions occurs in visible space which in general is hard to optimize. Based on this observation \cite{li2019aim} match the join distribution minimizing the divergence in latent space and the conditional in visible space, the decomposition is similar to VAE approach and can be observed in equation \ref{aim_match}. 
\begin{align}
    & D_{KL}(p(z, x) ||q(z, x)) \nonumber \\
    ~ = ~    &\mathbb{E}_{p(z)}[D_{KL}(p(x|z) ||q(x|z))] + D_{KL}(p(z) ||q(z)) \nonumber \\
    ~ = ~   &\mathbb{E}_{p(z)} \mathbb{E}_{p(x|z)}[  -  \log q(z|x)    ] + \mathbb{E}_{p(z)} [D_{KL}(p(x|z)|| q(x)) ]  + \mathbb{E}_{p(z)} [ \log p(z)],
    \label{aim_match}
\end{align}
%\ph{Add proof of this Eq to Appendix}
where $p(z)$ is a fixed parametric distribution hence $\mathbb{E}_{p(z)} [ \log p(z)]$ is constant. Therefore \cite{li2019aim} achieves the matching of joint distributions by minimizing \\ $D_{KL}(p(x|z)|| q(x))$ to learn the real domain, and maximizing the likelihood of the encoder $\mathbb{E}_{p(x|z)}[\log q(z|x)]$. This allows obtaining an overall better performance than \cite{dumoulin2016ali}, \cite{donahue2016bigan}, \cite{li2017alice} in terms of reconstruction and generation scores. %\ph{I would change this sentence} 
This method matches the conditional distribution explicitly, uses a flexible generator \cite{goodfellow2014gan} and avoids latent collapse problems \cite{Lucas2019UnderstandingPC}. 

Lot of research has been done in unsupervised and semi supervised learning using straight forward joint distribution optimization \cite{donahue2016bigan}, \cite{dumoulin2016ali}, \cite{bigbigan}, \cite{amm}, and even more for conditional in latent space decomposition \cite{kingma2013vae}, \cite{kingma2016dsemi}, \cite{maaloe16}, \cite{maale2019biva}. In this work we explore the representation capabilities of the decomposition proposed in \cite{li2019aim}. Our main contributions are:

\begin{itemize}
    \item A mathematical derivation that allows us to have a varied mixture of distributions in latent space enforcing its clustering capabilities. Based on this derivation we developed a new generative model for clustering called MPCC, trained by matching prior and conditional distributions jointly. %MPCC performs clustering while retaining inference capabilities and \ph{superior} good conditional sampling performance.
    \item A comparison with the state-of-the-art  showing that MPCC outperforms generative and discriminative models in terms of clustering accuracy and generation quality.
    \item  An ablation study of the most relevant parameters of MPCC and a comparison with the AIM baseline \cite{li2019aim} using state of the art architectures \cite{bigbigan}. %We note the relevance of a learnable prior and the number of encoder updates.%, exploring the relevance of the number encoder updates when using parameter sharing between the discriminator and the encoder.
\end{itemize}

%Lot of research has been done in unsupervised and semi supervised learning using straight forward joint distribution optimization \cite{donahue2016bigan}, \cite{dumoulin2016ali}, \cite{bigbigan}, \cite{amm}, and even more for conditional in latent space decomposition \cite{kingma2013vae}, \cite{kingma2016dsemi}, \cite{maaloe16}, \cite{maale2019biva}. In this work we focus in the clustering task, matching the conditionals in data space and priors in latent space. This method enjoys the advantages of match the conditional distribution explicitly instead of \cite{donahue2016bigan}, \cite{dumoulin2016ali}, \cite{bigbigan}, \cite{amm}, have a flexible generator \cite{goodfellow2014gan} and avoids latent collapse problems \cite{Lucas2019UnderstandingPC}.

\section{Method}
\label{method_section}

%\ph{Homework for tonight: Check inconsistencies}
%The framework proposed in \cite{li2019aim} matches the joint distributions $q(x,z)$ and $p(x,z)$ by matching the prior and conditionals explicitly. This model has the advantage of using
%Matching conditional and priors \cite{li2019aim} have appeared as a good methodology for learning joint distributions. This procedure enjoys both advantages, 
%an encoder to infer the latent variables from data and a flexible decoder that doesn't assume any parametric distribution for the generated output \ph{This section should describe our method not others, this sentence should not be here unless we use it to describe MPCC}\na{wich sentence?}.\pe{la oracion con que comienza esta unidad no esta bien porque es como una descripcion de otro metodo. En la seccion de metodos hay que ir directo a lo que se propose. Algo asi como. In this section we extend the graphical model presented in [16] by adding an additional latent variable, y, which is categorical and represents a given clustering }  

%In this section we extend the graphical model presented in \cite{li2019aim} by adding an additional latent variable, $y$, which represents a given clustering.

\subsection{Model definition}
\label{model_def}

MPCC extends the usual joint distribution of variables $x \in \mathcal{X}$ and $z \in \mathcal{Z}$ incorporating an additional latent variable, $y \in \mathcal{Y}$, which represents a given cluster.
%To make inference in a cluster or class an additional latent variable $y$ is required. We extend the graphical model presented in \cite{li2019aim} for that purpose, both 
%We specify the \textit{encoder} and \textit{decoder} joint distributions as
We specify the graphical models for \textit{generation} and \textit{inference} as

\begin{center}
\begin{minipage}{.5\textwidth}
\begin{itemize}
    \item %The \textit{decoder} joint distribution 
    $p(x,z,y) = p(y) p (z|y) p(x|z,y)$,
    \item %The \textit{encoder} joint distribution 
    $q(x,z,y) = q (y|z) q(z|x) q(x)$,
\end{itemize}
\end{minipage}
\end{center}
respectively. The only assumption in the graphical model is $q(y|z)=q(y|z,x)$, \textit{i.e.} $z$ contains all the information from $x$ that is necessary to estimate $y$. 

For generation, we seek to match the decoder $p(x|z, y)$ to the real data distribution $q(x)$. The latent variable is defined by the conditional distributions $p(z|y)$ which in general can be any distribution under certain restrictions (Section \ref{optimizing_mpcc}). %The training procedure will enforce that the latent variables $z|y_k$ are separable under different $y_k$. 
The marginal distribution $p(y)$ is defined as multinomial with weight probabilities $\phi$. Note that under this graphical model the latent space becomes multimodal defined by a mixture of distributions $p(z) = \sum_y p(y)p(z|y)$.

 %For generation, we seek to match the decoder $p(x|z, y)$ to the real data distribution $q(x)$. The latent variable $z|y \sim \mathcal{N}(\mu_y, \sigma^2 _y)$ it is assumed gaussian and is sampled using the reparameterization trick \cite{kingma2013vae}, \textit{i.e.} $z = \mu_{y}  +  \sigma_{y} \odot \epsilon$ where $\epsilon \sim \mathcal{N}(0,I)$ and $\odot$ is the Hadamard product. The parameters $\mu_{y}$, $\sigma^2_{y}$ are learnable and they are conditioned on $y$. The marginal distribution $p(y)$ is defined as multinomial with weight probabilities $\phi$. 
 
In the inference procedure the latent variables are obtained by the conditional posterior $q(z|x)$ using the empirical data distribution $q(x)$. The  distribution $q(y|z)$ is a posterior approximation  of the cluster membership of the data. %\ph{Should we state that we assume $q(y|z,x)=q(y|z)$?}\na{probably, but i'm not 100 \% surex.}\pe{Yes!, now it makes sense to me}

We call our model Matching Priors and Conditionals for Clustering (MPCC) and we optimize it by minimizing the reverse Kullback-Leibler divergence of the conditionals and priors between the inference and generative networks as follows:
\begin{align}
\begin{split}
    &D_{KL} \left ( p(x,z,y) || q(x, z, y) \right) \\ 
    ~ = ~ &\mathbb{E}_{p(z,y)}[D_{KL}(p(x|z,y) || q(x|z,y))]  \\
     + ~&\mathbb{E}_{p(y)}[D_{KL}(p(z|y) || q(z|y))] + D_{KL}(p(y) || q(y)) .
    \label{loss}
\end{split}
\end{align} 
The proof for Eq. \eqref{loss} can be found in Appendix A. In the following sections we derive a tractable expression for Eq. \eqref{loss} and present the MPCC algorithm. 

\subsection{Loss function}

Because $q(y)$, $q(z|y)$ and $q(x|z,y)$ are impossible to sample from, we derive a closed-form solution for Eq. \eqref{loss}. In particular for any fixed $y$ and $z$ we can decompose  $D_{KL}(p(x|z,y)||q(x|z,y))$ as follows:
\begin{align}
    & D_{KL}(p(x|z,y) || q(x|z,y)) \nonumber \\ ~ = ~ 
    & \mathbb{E}_{p(x | z,y)}\left[ \log \frac{p(x|z,y)}{q(x)} \frac{q(z,y)}{q(z,y|x)}  \right]  \nonumber\\  ~ = ~
    & \mathbb{E}_{p(x | z,y)}  \bigg[ \log \frac{p(x|z,y)}{q(x)} - \log q(y|z) - \log q(z|x) + \log q(z|y) + \log q(y)  \bigg]. 
    \label{decompose}
\end{align}
%To obtain the objective function \ref{loss} from the last equation two additional terms are required, 
Adding $\log p(z|y) + \log p(y)  - \log q(z|y)  - \log q(y) $  to both sides of Eq. \eqref{decompose} and taking the expectation with respect to $p(z,y)$ the  Eq. \eqref{loss} is recovered. After adding these terms and taking the expectation we can collect the resulting right hand side of Eq. \eqref{decompose} as follows: %By collecting terms in the right hand side Rearranging terms on the right hand side \pe{of Eq. \eqref{loss}} \ph{revisar} we can rewrite it as follows:
\begin{align}
    &\mathbb{E}_{p(z,y)}[D_{KL}(p(x|z,y) || q(x|z,y)) + D_{KL}(p(z|y) || q(z|y)) + D_{KL}(p(y) || q(y)) ]  \nonumber \\ 
     =  ~ & \underbrace{\mathbb{E}_{p(y)p(z|y)} [D_{KL}(p(x|z,y)||q(x))]}_{\textbf {Loss I} } + \underbrace{\mathbb{E}_{p(y)p(z|y)p(x|z,y)} [- \log q(z|x) - \log q(y|z)  ]}_{\textbf{Loss II} } \nonumber \\ 
     +~& \underbrace{\mathbb{E}_{p(z|y)p(y)}[\log p(y)+ \log p(z|y)]}_{\textbf{Loss III}}, \label{finalloss}
\end{align}

where \textbf{Loss I} seeks to match the true distribution $q(x)$, \textbf{Loss II} is related to the variational approximation of the latent variables and \textbf{Loss III} is associated with the distribution of the cluster parameters. The right hand term of Eq. \eqref{finalloss} is a loss function, composed of three terms with distributions that we can sample from. In the next section we explain the strategy to optimize each of the terms of the proposed loss function.

MPCC follows the idea that the data space $\mathcal{X}$ is compressed in the latent space $\mathcal{Z}$ and a separation in this space will likely partition the data in the most representatives clusters $p(z|y)$. The separability of these conditional distributions will be enforced by $q(y|z)$ which also backpropagates through the parameters of $p(z|y)$. The connection with the data space is through the decoder $p(x|z,y)$ for generation and the encoder $q(z|x)$ for inference.
%: $q(y|z)$ will guarantee the separability of all the conditional distributions $p(z|y)$. Each conditional $p(z|y)$ will represent a cluster in the compress space $z$
%and the connection with the data space is throught the decoder $p(x|z,y)$ for  generation and with $q(z|x)$ for inference.

\subsection{Optimizing MPCC}
\label{optimizing_mpcc}

In what follows we describe the assumptions made in the distributions of the graphical model and how to optimize Eq. \ref{finalloss}. For simplicity we assume the conditional $p(z|y)$ to be a Gaussian distribution, but other distributions could be used with the only restriction being that their entropy should have a closed-form or at least a bound (second term in \textbf{Loss III}). In our experiments the latent variable $z|y \sim \mathcal{N}(\mu_y, \sigma^2 _y)$ is sampled using the reparameterization trick \cite{kingma2013vae}, \textit{i.e.} $z = \mu_{y}  +  \sigma_{y} \odot \epsilon$ where $\epsilon \sim \mathcal{N}(0,I)$ and $\odot$ is the Hadamard product. The parameters $\mu_{y}$, $\sigma^2_{y}$ are learnable and they are conditioned on $y$. Under Gaussian conditional distribution the latent space becomes a GMM, as we can observe mathematically $p(z) = \sum_y p(y)p(z|y) = \sum_y p(y)\mathcal{N}(\mu_y, \sigma^2 _y)$. 

\begin{figure*}[t]
\begin{center}
    \includegraphics[width=0.95\textwidth]{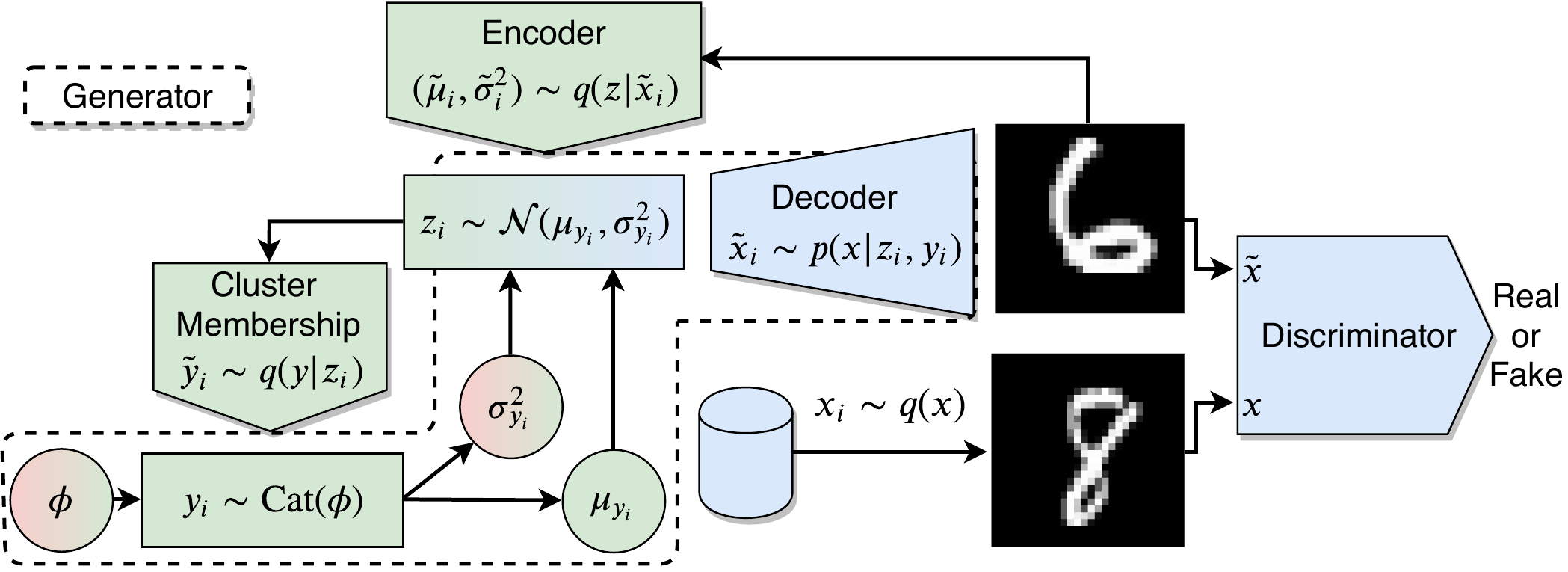}
\end{center}
      \caption{\label{fig-MPCC} Diagram of the MPCC model. The blue colored elements are associated with \textbf{Loss I} (Eq. \ref{GAN}). The green colored elements are associated with \textbf{Loss II} (Equations \ref{gaussian_gaussian} and \ref{classifier_loss}). The red colored elements are associated with \textbf{Loss III} (Eq. \ref{prior_eq}). The dashed line corresponds to the generator (GMM plus decoder).}
\end{figure*}

The distribution $p(x|z,y)$ is modeled by a neural network and trained via adversarial learning, \textit{i.e.} it does not require parametric assumptions. The inferential distribution $q(z|x)$ is also modeled by a neural network and its distribution is assumed Gaussian for simplicity. The categorical distribution $q(y|z)$ may also be modeled by a neural network but we propose a simpler approach based on the membership from the latent variable $z$ to the Gaussian components. A diagram of the proposed model considering these assumptions is shown in Fig. \ref{fig-MPCC}. We now expand on this for each of the losses in Eq. \ref{finalloss}.

%In principle any finite mixture of distributions can be used as prior for our model. The only restrictions are for the entropy of $p(z|y)$ to have a closed-form or at least a bound (second term in \textbf{Loss III}). As $p(x|z,y)$

%We will observe that any mixture distribution can be used that complies with certain restrictions. Under these assumptions the latent space becomes a GMM, but any mixture of distributions (Section \ref{model_def}) could be used under certain restrictions. 
 
%We estimate $q(z|x)$ and $p(x|z,y)$ using neural networks, and we consider $q(z|x)$ and $p(z|y)$ as gaussians. 
 
%$q(y|z)$ is a categorical distribution computed as the membressy of $z$ to each gaussian in the prior. 
 
%In general $q(z|x)$ and $p(z|y)$ can have any distribution, the only difficult restriction comes from the conditional distribution $p(z|y)$ in which its entropy should be calculable or at least has bound (second term in \textbf{Loss III}). 
 
%$q(y|z)$ it not limited by the change in $p(z|y)$ since all distributions in the \textit{inference} and \textit{generative} model can be computed through neural networks. 

%In whats left of the paper we will refer to MPCC to this case. 

\textbf{Loss I}: Instead of minimizing the Kullback-Leibler divergence shown in the first term on the right hand of Eq. \eqref{finalloss} we choose to match the conditional decoder $p(x|z,y)$ with the empirical data distribution $q(x)$ using a generative adversarial approach. %This allows us to avoid assuming a parametric distribution for the generator output. 
The GAN loss function can be formulated as \cite{Dong2019TowardsAD} %shown in Eq. \eqref{GAN} and one of the main advantages is that it doesn't assume any parameterization of the generated output, this is particularly useful when dealing with high dimensional data [GAN papers].
\begin{align}
\begin{split}
 & \max_{D}~ \mathbb{E}_{x\sim q(x)} [ f( D (x) )] + \mathbb{E}_{\tilde{x} \sim p(x,z,y)} [  g( D (\tilde{x})) ],\\
 & \min_{G}~ \mathbb{E}_{\tilde{x} \sim p(x,z,y)} [  h( D (\tilde{x})) ],
\label{GAN}
\end{split}
\end{align}
%\na{We should explain the sampling of p(x,z,y) y tried to put hinge loss the simplest way possible, with the montecarlo samplig but the space is not that much. We also need to explain hinge loss with the function that i defined https://arxiv.org/pdf/1901.08753.pdf}
%where $D$ and $G$ are the discriminator and generator networks, respectively. The parameters and distribution associated with \textbf{Loss I} are colored in blue in Fig. \ref{fig-MPCC}.
where $D$ and $G$ are the discriminator and generator networks, respectively, and tilde is used to denote sampled variables. For all our experiments we use the hinge loss function \cite{lim2017geometric}, \cite{tran2017hierarchical}, \textit{i.e.} $f = -\min(0, o -1)$, $g = \min(0, - o - 1)$ and $h = -o$, being $o$ the output of the discriminator. The parameters and distribution associated with \textbf{Loss I} are colored in blue in Fig.\ref{fig-MPCC}.

%Notice that minimize \textbf{Loss II} is equivalent to reduce the cross-entropy between the prior and inference distribution. 

\textbf{Loss II}: The first term of this loss is %equivalent to the cross-entropy between the prior and inference distribution and we express it as 
estimated through Monte Carlo sampling as
\begin{align}
\begin{split}
       &\mathbb{E}_{p(y)p(z|y)p(x|z,y)}[- \log q(z|x) ]\\ = ~
       &\mathbb{E}_{ y_i \sim p(y),  z_i \sim p(z|y = y_i),  \tilde{x}_i \sim p(x|z = z_i,y = y_i)} \underbrace{ \left [\sum_{j=1}^J \frac{1}{2}\log (2\pi \tilde{\sigma}_{ij}^2) +  \frac{(z_{ij} - \tilde{\mu}_{ij} )^2}{2 \tilde{\sigma} _{ij}^2}  \right]}_{L_q(\tilde{\mu_i}, \tilde{\sigma}^2_i, z_i)} ,
      \label{gaussian_gaussian}
\end{split}
\end{align}
where $J$ is the dimensionality of the latent variable $z$. By minimizing Eq. \eqref{gaussian_gaussian} we are %Eq. \eqref{gaussian_gaussian} is optimized by 
maximizing the log-likelihood of the encoder $q(z|x)$ with respect to the Gaussian prior $p(z|y)$. This reconstruction error is estimated %through monte carlo 
by matching the samples $z_i \sim p(z|y=y_i)$ with the Gaussian distribution $(\tilde{\mu}_i, \tilde{\sigma}^2_i )\sim q(z|x=\tilde{x}_i)$, where $\tilde{x}_i$ is the decoded representation of $z_i$. %\pe{define J in eq 7}

The second term of \textbf{Loss II} is equivalent to the cross-entropy between the sampled label $y_i \sim p(y)$ and the estimated cluster membership $\tilde{y}_i$

\begin{align}
    L_c (y_i, \tilde{y}_i) = -\sum_{k=1}^K y_{ik} \log \tilde{y}_{ik},
    \label{classifier_loss}
\end{align}
where $K$ is the number of clusters and %\ph{missing $\phi_k$}
\begin{align}
    \tilde{y}_{im} = q(y = m|z = z_i) = \frac{ \mathcal{N}(z_i| \mu_{m}, \sigma_m^2) } {\sum_{k=1}^K \mathcal{N}(z_i| \mu_{k}, \sigma_k^2)  },
    \label{gaussian_membressy}
\end{align}
is the membership of $z_i$ to the m-th cluster. The parameters $\mu_m$ and $\sigma^2_m$ are learnable, and $m \in [1, \dots, K]$ is the index corresponding to each cluster. The parameters and distribution associated with \textbf{Loss II} are colored in green in Fig. \ref{fig-MPCC}. In practice Eq. \ref{gaussian_membressy} is estimated using the log-sum-exp trick. %Note that the loss in Eq. \ref{classifier_loss} backpropagates through the parameters of $q(y|z)$ and $p(z|y)$ enforcing the separation of the latent variables, which is fundamental for the clustering capabilities of MPCC. %Note that in Eq. \ref{gaussian_membressy} a neural network could be used to estimate the posterior of the categorical distribution  allowing to more flexible distribution, we used gaussian membressy for simplicity.\na{m}

\textbf{Loss III}: This loss is associated with the regularization of the Gaussian mixture model parameters $\phi$, $\mu$ and $\sigma^2$ and has a closed form 
\begin{align}
\begin{split}
    \label{prior_eq}
    & \mathbb{E}_{p(y)p(z|y)}[ \log p(y) + \log p(z|y) ] \\ = & \underbrace{\sum_{k=1}^K \phi_{k} \left[ \log \phi_{k}  - \sum_{j=1}^J  \left(\frac{1}{2} + \frac{1}{2}\log (2\pi \sigma_{kj}^2) \right)\right] }_{L_p(\phi, \sigma^2)},
\end{split}
\end{align}

where the first term corresponds to the entropy maximization of the mixture weights, \textit{i.e.} in general every Gaussian will not collapse to less than K modes of the data distribution which is a  solution with lower entropy. In our experiments we fix $\phi_k =1/K$, \textit{i.e.} $\phi$ is not learnable. The second term is a regularization for the variance (entropy) of each Gaussian which avoids the collapse of $p(z|y)$. The parameters associated with \textbf{Loss III} are shown in red in Fig. \ref{fig-MPCC}.

\textbf{Loss I} scale differs from that of the terms associated with the latent variables. To balance all terms we multiply Eq. \eqref{gaussian_gaussian} by one over the dimensionality of $x$\footnote{If x is an image then its dimensionality would be $channels \times height \times width$} and the second term of Eq. \eqref{prior_eq} by one over the dimensionality of the latent variables. During training \textbf{Loss III} is weighted by a constant factor $\lambda_p$. We explain how this constant is set in Section \ref{empirical_setup}. The full procedure to train the MPCC model is summarized in Algorithm \ref{mpcc_alg}. Note that MPCC is scalable in the number of clusters since Eq. \ref{gaussian_gaussian} is a Monte Carlo approximation in $y$ and the cost of Eq. \ref{prior_eq} is low since $J$ is small in comparison to the data dimensionality.

\begin{algorithm}[ht]
\caption{MPCC algorithm}\label{mpcc_alg}
\setstretch{0.4}
\begin{algorithmic}[1]
\State $K$, $J \gets$ Set number of clusters and latent dimensionality 
\State $\eta$, $\eta_p  \gets$ Set learning rates 
\State $\theta_g, \theta_d, \theta_e \gets$ Initialize network parameters
\State $\phi, \mu, \sigma^2 \gets $ Initialize GMM parameters
\State $\theta_c \gets [\phi, \mu, \sigma^2 ]$
\Repeat
\For{$D_{steps}$}
    \State $x_1, \dots, x_n \sim q(x)$
    \Comment{Draw n samples from empirical distribution}
    \State $y_1, \dots, y_n \sim p(y)$ 
    \Comment{Draw n samples from categorical prior}
    \State $z_i \sim p(z|y = y_i),  ~~~~~ i =1   ,\dots, n$ 
    \Comment{Draw n samples from Gaussian conditional prior}
    \State $\tilde{x}_i \sim p(x|z = z_i,y =  y_i),  ~~~~~ i =1   ,\dots, n$ 
    \Comment{Generate samples using generator network}
    \State $\theta_d \gets \theta_d + \eta \nabla_{\theta_d}\left[\frac{1}{n}\sum_{j=1}^n f( D(x_j) ) +     \frac{1}{n}\sum_{i=1}^n g(D(\tilde{x}_i) ) \right]$
\Comment{Gradient update on discriminator network}
\EndFor
\State $y_1, \dots, y_n \sim p(y)$ 
\Comment{Draw n samples from categorical prior}
\State $z_i \sim p(z|y = y_i),  ~~~~~ i =1   ,\dots, n$ 
\Comment{Draw n samples from Gaussian conditional prior}
\State $\tilde{x}_i \sim p(x|z = z_i,y =  y_i),  ~~~~~ i =1   ,\dots, n$ 
\Comment{Generate samples using generator network}
\State $(\theta_g, \theta_c )\gets (\theta_g, \theta_c ) - \eta \nabla_{(\theta_g, \theta_c )} \frac{1}{n}\sum_{i=1}^n h( D(\tilde{x}_i) ) $
\Comment{Gradient update on generator network}
\For{$E_{steps}$}
    \State $y_1, \dots, y_n \sim p(y)$ 
    \Comment{Draw n samples from categorical prior}
    \State $z_i \sim p(z|y = y_i),  ~~~~~ i =1   ,\dots, n$ 
    \Comment{Draw n samples from Gaussian conditional prior}
    \State $\tilde{x}_i \sim p(x|z = z_i,y =  y_i),  ~~~~~ i =1   ,\dots, n$ 
    \Comment{Generate samples using generator network}
    \State $(\tilde{\mu}_i, \tilde{\sigma}^2_i) \sim q(z| x = \tilde{x}_i),  ~~~~~ i =1   ,\dots, n$ 
    \Comment{Encode $\tilde{x}$ to obtain mean and variance}
    \State $ \theta_e \gets \theta_e - \eta \nabla_{\theta_e}\frac{1}{n}\sum_{i=1}^n L_q(\tilde{\mu}_i,\tilde{\sigma}^2_i, z_i)$
    \Comment{Gradient update on encoder network}
    \If{first $E_{step}$} 
    \State $\tilde{y}_i \sim q(y|z = z_i),  ~~~~~ i =1   ,\dots, n$ 
    \State  $\theta_c  \gets  \theta_c - \eta_p \nabla_{\theta_c} \left[ \frac{1}{n}\sum_{i=1}^n L_c (y_i, \tilde{y}_i) + \lambda_p \cdot L_p(\phi, \sigma^2)) \right]$
    \Comment{Gradient update on Prior parameters}
    \EndIf
\EndFor
\Until{convergence}
\end{algorithmic}
\end{algorithm}
%\vspace{-5mm}

\section{Related methods}
\label{literature_review}

In Section \ref{method_section} we showed that the latent space of MPCC it is reduced to a GMM under Gaussian conditional distribution. Because all the experiments are performed based on this assumption in this section we summarize the literature of generative and autoencoding models that consider GMMs.

The combination of generative models and GMMs is not new. Several methods have applied GMM in autoencoding \cite{AGGMM}, \cite{AnomalyGMM} or GAN \cite{DeliGAN}, \cite{OGANSGMM} applications without clustering purposes. Other approaches have performed clustering but are not directly comparable since they use mixtures of various generators and discriminators \cite{MGANC} or  fixed priors with ad-hoc set parameters \cite{GAN_ICLR}.

%GMM has been applied in autoencoding applications for language processing \cite{AGGMM} and anomaly detection \cite{AnomalyGMM}, and in GANs to overcome low data scenarios \cite{DeliGAN} and for comparison purposes in data space \cite{OGANSGMM}. These techniques consider GMM but they do not perform clustering and thus are not directly comparable to our work. Fixed GMM priors with ad-hoc set parameters are not comparable either \cite{GAN_ICLR} as this assumption limits their applications in unsupervised settings. 
%GMM clustering methods using GANs has been studied using mixture of various generators and discriminators \cite{MGANC}, and in VAEs with GMM priors \cite{jiang2017vade}, \cite{GVADE_ICCV}.

%In \cite{OGANSGMM} they fit GMM in visible space for comparison purposes, in \cite{AnomalyGMM} DAGMM estimated gaussian parameter for anomaly detection, in \cite{DeliGAN} DeliGAN explore GMM in the prior to deal with low data scenarios. Note that the techniques mentioned consider GMM in their model but do not performed clustering and thus are not comparable.

%\ph{Nico, check that what I wrote here and in 3.2 is consistent}
Among the related works on generative models for clustering the closest approaches to our proposal are ClusterGAN \cite{Mukherjee2019ClusterGANL} and Variational Deep Embedding \cite{jiang2017vade}. %The main difference between MPCC and ClusterGAN architecture \na{revisar} being on how the embedding is made to do clustering. While 
ClusterGAN differs from our model in that it sets the dimensions of the latent space as either continuous or categorical while MPCC uses a continuous latent space which is conditioned on the categorical variable $y$. %\ph{the advantage of the latter is that we can the conditional sampling?}. 
On the other hand, Variational Deep Embedding (VADE) differs greatly in the training procedure, despite its similar theoretical basis. VADE, as a variational autoencoder model, matches the joint distributions in the forward KL sense $D_{KL}(q(x,z,y)||p(x,z,y))$ by matching the posteriors and the marginals in data space %\ph{sometimes we say visible, other times we say data}
as demonstrated in Appendix B% included, as this is not proven in \cite{jiang2017vade} we included the demonstration in the supplementary material
. MPCC optimizes the reverse KL, \textit{i.e.} matching the priors in latent space and conditionals in data space. Optimizing different KLs yield notably different decompositions and thus training procedures. For the forward KL \cite{jiang2017vade} in addition to the challenges in scaling to larger dimension (Section \ref{background}) it is more difficult to generalize the latent space to any multi-modal distribution, we briefly discuss the reasons for this in Appendix B.  %\ph{Si q es la real y p es la aproximacion DKL(q||p) es la forward, pero en VAE/VADE usualmente q es la aproximación, esto podria inducir a confusion}

\section{Experiments}

\subsection{Quantitative Comparison}
Following \cite{DEC}, the performance of MPCC is measured using the clustering accuracy metric in which each cluster is assigned to the most frequent class in the cluster. %\ph{Is this equivalent to the "best mapping" stuff you mention below?}\na{yep}. 
%\begin{comment}
Formally this is defined as 
\begin{equation}
    \text{ACC} = \max_{m \in \mathcal{M}}\frac{\sum_{i = 1}^N \mathds{1} \{ y_i = m (c_i) \} }{N},
    \label{purity}
\end{equation}
where $N$ is the total number of samples, $y_i$ is the ground truth, $c_i = \text{arg}\max_k q(y=k|z=z_i)$ is the predicted cluster %\ph{In this case $\text{arg}\max_k q(y=k, z=z_i)$ right?} 
and $\mathcal{M}$ is the space of all possible mappings between clusters and labels.
%\end{comment}

To measure the quality of the samples generated by MPCC we use the inception score (IS) \cite{improveTechGAN} and the Fr\'echet inception distance (FID) \cite{FID}.

\subsection{Datasets}
\label{sec_dataset}

In order to evaluate MPCC we performed clustering in five benchmark datasets: a handwritten digit dataset (MNIST, \cite{lecun1998mnist}), a handwritten character dataset (Omniglot, \cite{lake2015omniglot}), two color image dataset (CIFAR-10 and CIFAR-100 \cite{krizhevsky2009cifar}) and a fashion products image dataset (Fashion-MNIST, \cite{xiao2017fashion}). For CIFAR-100 we consider the 20 superclasses. Omniglot was created using the procedure described in  \cite{imsat}. Because the task is fully unsupervised we concatenate the training and test sets as frequently done in the area \cite{imsat}, \cite{DAC}, \cite{DEC}. All datasets have 10 classes except for Omniglot  and CIFAR-20 with 100 and 20 respectively. All images were resized to $32 \times 32$ and reescaled to [-1,1] in order to use similar architectures. The CIFAR-10 experiments where IS and FID are reported (tables \ref{ablation_tab}, \ref{comparation_tab} and \ref{metric_tab}) were trained using only the training set (50,000 examples) for a fair comparison with the literature. For all clustering experiments we use the same number of cluster as the datasets classes. %\ph{future work, analizar que pasa cuando esto no se cumple}
%Notice that models that give IS or FID scores usually use just the training set, for this reason CIFAR-10 experiments of tables \ref{ablation_tab}, \ref{comparation_tab} and \ref{metric_tab} were trained with 50,000 points. 
%\ph{what was this?}\na{this procedure is just for Omniglot, and is data augmentation like, scaling, translation, rotation.}. 
%Reuters was obtained from (VADE github) which contains 2000 tf-idf features from 2000 most frequent word to represent articles. A summary of all dataset statistics can be found in Table \ref{data_tab}

\begin{comment}
\begin{table}[t]
  \centering
      \caption{Dataset statistics}
  \scalebox{0.9}{
  \begin{tabular}{lccccc}
    \toprule    
    Datasets     & \# Examples   & \# Classes & Dimension \\
    \midrule
    MNIST    & 70000  & 10 &  1$\times$28$\times$28  \\
    Fashion-MNIST & 70000  & 10 &  1$\times$28$\times$28      \\
    Omniglot& 40000  & 100 & 1$\times$32$\times$32    \\
    CIFAR-10 & 60000  & 10 &  3$\times$32$\times$32     \\
    CIFAR-20 & 60000  & 20 &  3$\times$32$\times$32     \\
    %Reuters & 10000  & 4 &  2000  & 43\%    \\
    \bottomrule
  \end{tabular}}
  \label{data_tab}
  %\vspace{-5mm}
\end{table}
\end{comment}

\subsection{Empirical details}
\label{empirical_setup}

Our architecture is based on optimization techniques used in the BigGAN  \cite{brock2018large}\footnote{\url{https://github.com/ajbrock/BigGAN-PyTorch}}, we found that simpler architectures such as DCGAN \cite{DCGAN} were not able to learn complex distributions like CIFAR-10 while optimizing the parameters of the prior. %In Appendix C  all the architecture details can be found. 
Architecture details are given in Appendix C. We consider parameter sharing between the encoder and discriminator, and we test the importance of this in Section \ref{comparation study}. We set $D_{steps} = 4$ (see Algorithm \ref{mpcc_alg}), 
%we found empirically that in CIFAR-10 for smaller values of $D_{steps}$ mode collapse problems would appear during the early stages of training (see Appendix D).
we found that using smaller values of $D_{steps}$ causes mode collapse problems when training on CIFAR-10 (see Appendix D). A similar effect can be observed when choosing a low number of latent dimensions, therefore we set $J = 128$ in all CIFAR-10 experiments. % For clustering experiments (Section \ref{clustering experiment})
We made small changes in the architecture and optimization parameters depending on the dataset (see Appendix C).

We observed the same relation between batch size and (IS, FID) reported in \cite{brock2018large}. However we found artifacts that hurt accuracy performance when using batch size larger than 50. For simplicity we used this value in all experiments. %\ph{otra cosa para el future work}

We consider a weighting factor $\lambda_p$ for \textbf{Loss III} (Eq. \ref{prior_eq}). We observed that if $\lambda_p=1$, the standard deviation of the prior $\sigma$ would increase monotonically, hindering training.  
%We found that if we optimize Eq. \eqref{finalloss} without weighting Eq. \eqref{prior_eq} appropriately the $\sigma$ tends to increase monotonically, hindering the training. \ph{I had to go back to the algorithm to see what lamdbap was} 
On the other hand if $\lambda_p$ is too small, $\sigma$ decreases, collapsing at some point. %To fix this we multiply Eq. \eqref{prior_eq} by a constant $\lambda_p = 0.01$ and set a minimum threshold for the standard deviation of the prior distribution of $0.5$. 
We found empirically that a value of $\lambda_p=0.01$ combined with a minimum threshold for $\sigma$ of $0.5$ allow the algorithm to converge to good solutions.

The parameter settings indicated above were fixed for all experiments and didn't show a big effect in accuracy performance. In section \ref{ablation study} we explore the parameters that most affect the training. We trained all experiments for 75,000 iterations, except for MNIST and Omniglot which iterate for 125,000. %\ph{MNIST más epocas que CIFAR10?, como se explica eso?}. 
For unconditional and conditional training we kept the model of the last iteration.

\subsection{Ablation study}
\label{ablation study}
We found that $E_{steps}$, the number of encoder updates per epoch, and $\eta_p$, the learning rate of the prior parameters, are the most relevant hyperparameters to obtain % good accuracy performance and
high accuracy and generation quality. Increasing $E_{steps}$ improves the estimation of $q(z|x)$ since the prior and generator parameters are changing constantly. Rows 1-3 of Table \ref{ablation_tab} show that the reconstruction error (MSE) decreases with $E_{steps}$.  Generation quality metrics (IS, FID) also improve with larger values of $E_{steps}$ due to the shared parameters between encoder and discriminator. 

At initialization the GMM components might not be separated. We observed that the clustering accuracy drops when the generators learns a good approximation of the real distribution before the clusters are separated. To avoid this we use a larger learning rate for the parameters of the GMM prior with respect to the parameters of the generator, encoder and discriminator. Rows 4-6 of Table \ref{ablation_tab} show that the clustering accuracy increases for larger values of $\eta_p$.

\begin{table}[t]
\begin{minipage}[t]{0.48\linewidth}
%\ph{se pueden hacer horizontales para que se lean mejor los $\pm$}
\centering
\caption{$E_{steps}$ correspond to the encoder updates and $\eta_p$ to the learning rate of the prior parameters. The scale of MSE is in $10^{-3}$. The statistics were obtained for at least three runs }
\scalebox{0.87}{%
  \begin{tabular}{ll|cccc}
    \toprule 
    $E_{steps}$ &  $\eta_p$     & Acc \%   & IS & FID & MSE \\
    \midrule
    1  &  2e-4   & \spc[t]{$41.31$  \\ $\pm 5.74$}  & \spc[t]{$8.82$  \\ $\pm 0.07$}   & \spc[t]{$11.38$  \\ $\pm 0.23$} & \spc[t]{$1.34$  \\ $\pm 0.96$} \\
    2   & 2e-4  & \spc[t]{$38.67$  \\ $\pm 3.52$}  & \spc[t]{$9.02$  \\ $\pm 0.05$}  & \spc[t]{$9.66$  \\ $\pm 3.98$}  & \spc[t]{$1.01$  \\ $\pm 1.11$}   \\
    4   & 2e-4  &  \spc[t]{$38.27$  \\ $\pm 2.46$}  & \spc[t]{$9.25$  \\ $\pm 0.09$}  & \spc[t]{$7.50$  \\ $\pm 0.43$}  & \spc[t]{$0.331$  \\ $\pm 0.09$}    \\
    4   & 4e-4  &  \spc[t]{$52.58$  \\ $\pm 5.30$}  & \spc[t]{$9.44$  \\ $\pm 0.06$}  & \spc[t]{$6.55$  \\ $\pm 0.33$}  & \spc[t]{$0.48$  \\ $\pm 0.22$}    \\
    4   & 6e-4  &  \spc[t]{$61.99$  \\ $\pm 4.96$}  & \spc[t]{$9.49$  \\ $\pm 0.15$}  & \spc[t]{$6.59$  \\ $\pm 0.45$}  & \spc[t]{$1.04$  \\ $\pm 1.03$}     \\
    \bottomrule
  \end{tabular}}
  \label{ablation_tab}
  
  \end{minipage}
%\vspace{-3mm}}
\hfill
\begin{minipage}[t]{0.48\linewidth}
\centering
\caption{Comparison of MPCC and AIM-MPCC methods with sharing parameters (S) and without sharing (NS) on the CIFAR-10 dataset. The scale of MSE is in $10^{-3}$. The statistics were obtained for five runs}
\scalebox{0.87}{%
    \begin{tabular}{lcccc}
    \toprule    
    Model     & Acc \%   & IS & FID & MSE \\
    \midrule
    AIM-MPCC (NS)   & -  & \spc[t]{$8.24$  \\ $\pm 0.07$}  & \spc[t]{$21.55$  \\ $\pm 1.47$}  & \spc[t]{$1.52$  \\ $\pm 0.84$}  \\
    AIM-MPCC (S)& -  & \spc[t]{$9.09$  \\ $\pm 0.04$}  & \spc[t]{$10.42$  \\ $\pm 0.36$}  & \spc[t]{$1.64$  \\ $\pm 1.42$}\\
    %MPCC (NoS) &  \spc[t]{$48.36$  \\ $\pm 4.45$}  & \spc[t]{$7.65$  \\ $\pm 0.62$}  & \spc[t]{$41.67$  \\ $\pm 11.43$}  & \spc[t]{$2.02$  \\ $\pm 1.03$}       \\
    MPCC (S)  & \spc[t]{$61.99$  \\ $\pm 4.96$}  & \spc[t]{$9.49$  \\ $\pm 0.15$}  & \spc[t]{$6.59$  \\ $\pm 0.45$}  & \spc[t]{$1.04$  \\ $\pm 1.03$}    \\
    \bottomrule
  \end{tabular}}
  \label{comparation_tab}
%\vspace{-5mm}
\end{minipage}
\end{table}

\subsection{Comparison between GMM Prior and Normal Prior}
\label{comparation study}
%\pe{Niko: por alguna combinacion de teclas se me borro esta seccion y la tuve que re-escribir. Favor de revisar. En particular la U italica de la distribucion uniforme.}
Using the best configuration found in the ablation study, we performed a comparison %study 
with AIM \cite{li2019aim}, whose results are shown in Table \ref{comparation_tab}. 
We can consider AIM as a particular case of MPCC where a standard Normal prior is used instead of the GMM prior. %, although 
AIM does not perform clustering therefore we compare it with MPCC in terms of reconstruction and generation quality. We use the same architecture and parameter settings of MPCC for AIM and we denote this model as AIM-MPCC. %In Table \ref{comparation_tab} is refered as AIM-MPCC noting the same parameter configuration except for the prior. 
%\sout{Although AIM does not perform clustering, we can compare with MPCC in terms of reconstruction and generation quality.} 
To extend our analysis further, Table \ref{comparation_tab} includes the results of using parameter sharing between the encoder and the discriminator (%see architecture details in 
Appendix C), an idea that was considered but not fully explored in % \pe{what do you mean by briefly observed?} \ph{mentioned but not fully-explored by} briefly observed in 
\cite{li2019aim}. %AIM is a particular case of MPCC\pe{Check this!}, when using prior distribution $U(-1,1)$ instead of GMM, and optimizing Eq. \ref{GAN} of \text{Loss I} and Eq. \ref{gaussian_gaussian} of \text{Loss II}. 

%The MPCC model with standard Normal prior and using the network architectures and optimization techniques described in Section \ref{empirical_setup} is called AIM-MPCC. %to differentiate the use of new optimization techniques from previous works\pe{Not clear, what do  you mean with the previous sentence? }. 
%To extend our analysis further, Table \ref{comparation_tab} includes the results of using parameter sharing between the encoder and the discriminator (see architecture details in Appendix \ph{3}), an idea that was mentioned but not explored in % \pe{what do you mean by briefly observed?} \ph{mentioned but not fully-explored by} briefly observed in 
%\cite{li2019aim}.

Note that AIM-MPCC (NS) is considered a baseline because the prior is Gaussian and the encoder doesn’t share parameters with the discriminator thus the existence of the encoder doesn’t affect the generation quality. %The performance training with and without an encoder in this case is exactly the same. 
In Table \ref{comparation_tab} we can observe the relevance of parameter sharing, with this conﬁguration ($E_{steps} = 4$) the baseline improves by 0.85 (IS) and 11.13 (FID) points. Adding the GMM in the prior improves an additional 0.4 (IS) and 3.93 (FID) points. In total when using the GMM Prior and parameter sharing with additional encoder updates we improve the baseline from 21.55 to 6.59 (69.4$\%$ improvement) in terms of FID score and 1.25 points (15.2$\%$ improvement)  in terms of IS. It is important to notice that these techniques are general and easily applied to any GAN scheme.%the baseline improves by 0.85 and 11.13 points in IS and FID scores respectively. Adding the GMM in the prior improves an additional 0.4 (IS) and 3.93 (FID) points. In total when using the GMM Prior and parameter sharing with additional encoder updates we improve the baseline from 21.55 to 6.59 (69.4$\%$ improvement) in terms of FID score and 1.25 points in terms of IS. It is important to notice that these techniques are general and easily applied to any GAN scheme.

%\end{comment}

\begin{comment}

\begin{table}[ht]
\centering
\caption{Comparison of MPCC and AIM-MPCC methods with sharing parameters (S) and without sharing (NS) on the CIFAR-10 dataset. The scale of MSE is in $10^{-3}$. The statistics were obtained for five runs.}
\scalebox{0.9}{
  \begin{tabular}{lcccc}
    \toprule    
    Model     & Acc \%   & IS & FID & MSE \\
    \midrule
    AIM-MPCC (NS)   & -  & $8.24 \pm 0.07$  & $21.55 \pm 1.47$  & $1.52 \pm 0.84$ \\
    AIM-MPCC (S)& -  & $9.09 \pm 0.04$  & $10.42 \pm 0.36$  & $1.64 \pm 1.42$\\
    %MPCC (NoS) &  \spc[t]{$48.36$  \\ $\pm 4.45$}  & \spc[t]{$7.65$  \\ $\pm 0.62$}  & \spc[t]{$41.67$  \\ $\pm 11.43$}  & \spc[t]{$2.02$  \\ $\pm 1.03$}       \\
    MPCC (S)  & $61.99 \pm 4.96$  & $9.49 \pm 0.15$  & $~6.59 \pm 0.45$  & $1.04 \pm 1.03$   \\
    \bottomrule
  \end{tabular}}
  \label{comparation_tab}
%\vspace{-10mm}

\end{table}
\end{comment}

\subsection{Generation quality of MPCC}
\label{generation quality}
%\ph{move this after clustering or change the order of the tables}

%From sections \ref{ablation study} and \ref{comparation study} we observe competitive results in terms of IS and FID, in table \ref{metric_tab} we compared MPCC with state of art methods using the configuration of row four from table \ref{comparation_tab}.

Using the configuration of row five from Table \ref{ablation_tab} we compare MPCC with nine state of the art methods, %in terms of generation quality metrics 
surpassing them in terms of IS and FID scores in both the unsupervised and supervised setting,
as shown in Table \ref{metric_tab}. %Overall we achieved state of the art results using IS and FID scores, in both the unsupervised and supervised settings. 
The unconditional generation is the most significant with an improvement of $46.9 \%$ (FID) over state-of-the-art (SOTA), AutoGAN \cite{AutoGANGong_2019_ICCV}. Most notably its performance is better than the current best conditional method (BigGAN). %Most notably the performance of the unconditional MPCC is better than the current best conditional method (BigGAN). 

%Figure \ref{fig-cond} shows sampled images for the conditional MPCC model where each column corresponds to a different cluster.  %\na{r} The quality metrics (IS, FID) of conditional training for five runs are $9.55 \pm 0.08$ (IS) and $5.69 \pm 0.17$ (FID).

\begin{table}[t]
  \centering
    \caption{Inception and FID scores for CIFAR-10, in unconditional and conditional training. Higher IS is better. Lower FID is better. $^{\dagger}$: Average of 10 runs.  $^{\ddagger}$: Best of many runs. $^{\dagger\dagger}$: Average of 5 runs. Results without symbols are not specified }
  \subfloat[Unconditional (unsupervised) generation]{
    \scalebox{0.9}{
  \begin{tabular}{lcc}
    \toprule
    Model & IS & FID \\
    \midrule
    DCGAN \cite{DCGAN} &  $6.64 \pm 0.14$  & $-$      \\
    SN-GAN$^{\dagger}$ \cite{miyato2018spectral} &  $8.22 \pm 0.05$  & $21.7\pm 2.1$    \\
    AutoGAN \cite{AutoGANGong_2019_ICCV} &  $8.55 \pm 0.10$  & $12.42$    \\
    PG-GAN  $^{\ddagger}$ \cite{karras2018progressive} &  $8.80 \pm 0.05$  & $-$     \\
    NCSN \cite{song2019generativeNCSN} &  $8.91$  & $25.32$     \\
    %DEPICT  & $96.3 $  & $-$ &  $-$  & $-$   & $-$  \\
    %MPCC (best) &  $9.67 \pm 0.076$  & $6.42$     \\
    \textbf{MPCC}$^{\dagger\dagger}$  &  $9.49 \pm 0.15$ & $6.59 \pm 0.45$ \\ 
    \bottomrule
    \end{tabular}}}
    \subfloat[Conditional (supervised) generation]{
      \scalebox{0.9}{
  \begin{tabular}{lcc}
    \toprule
    Model & IS & FID \\
    \midrule
    WGAN-GP \cite{IWAN} &  $8.42 \pm 0.10$  & $-$     \\
    %DEPICT  & $96.3 $  & $-$ &  $-$  & $-$   & $-$  \\
    SN-GAN $^{\dagger}$ \cite{miyato2018spectral}   &  $8.60\pm 0.08$ & $17.5$     \\
    Splitting GAN  $^{\ddagger}$ \cite{grinblat2017classsplitting} & $8.87 \pm 0.09$ & $-$     \\
    CA-GAN  $^{\dagger}$ \cite{CAGAN}        & $9.17 \pm 0.13$ & $-$     \\
    %DEPICT  & $96.3 $  & $-$ &  $-$  & $-$   & $-$  \\
     BigGAN \cite{brock2018large} &  $9.22$  & $-$     \\
     %MPCC (best) &  $9.62 \pm  0.13$  & $5.71$     \\
     \textbf{MPCC}$^{\dagger\dagger}$  &  $9.55 \pm  0.08$  & $5.69 \pm 0.17$     \\
    %DEPICT  & $96.3 $  & $-$ &  $-$  & $-$   & $-$  \\
    \bottomrule
  \end{tabular} }}

  \label{metric_tab}
  %\vspace{-5mm}
  
\end{table}

\subsection{Clustering experiments}
\label{clustering experiment}
%\ph{table and paragraph not yet updated}

Table \ref{cluster_tab} shows the clustering results for the selected benchmarks. We observe that in all the available benchmarks MPCC outperform the related methods, VADE \cite{jiang2017vade} and ClusterGAN \cite{Mukherjee2019ClusterGANL}.  In more complex datasets such as CIFAR10, MPCC notably surpass discriminative based models (\textit{e.g.} \cite{imsat}, \cite{ji2019invariant}) which are the most competitive methods in the current literature. For benchmarks with more classes the margin is even larger obtaining improvements over the SOTA of $\sim 42\%$ and $\sim 9.7 \%$ points in Omniglot and CIFAR-20 respectively, demonstrating empirically the scalability of MPCC when using a high number of clusters. %It can be observed that for all datasets our proposed method either achieves or surpasses the SOTA in terms of clustering accuracy. %\na{r} %Using the models with the best accuracy performance in figure \ref{fig-uns_samples} we show various samples for CIFAR-10 and MNIST dataset, in figure \ref{fig-uns_rec} we provide reconstructions using the original data.

%Table \ref{cluster_tab} \pe{POr alguna razon la tabla 4 aparece mencionada en el texto mucho despues que la tabla 5} shows the clustering results for the selected benchmarks. All the results of CIFAR-20 dataset were extracted from \cite{ji2019invariant}, the results of IMSAT and DEC from \cite{imsat}, the results of InfoGAN and ClusterGAN from \cite{Mukherjee2019ClusterGANL}  and the remaining from their respective papers. It can be observed that for all datasets our proposed method either achieves or surpasses the state-of-the-art  in terms of clustering accuracy. %\na{r} %Using the models with the best accuracy performance in figure \ref{fig-uns_samples} we show various samples for CIFAR-10 and MNIST dataset, in figure \ref{fig-uns_rec} we provide reconstructions using the original data.
 It can be observed that for all datasets our proposed method either achieves or surpasses the SOTA in terms of clustering. Figures \ref{fig-uns_samples} and \ref{fig-uns_rec} show examples of generated and reconstructed images, respectively, using the MPCC model with the highest accuracy in the MNIST and CIFAR-10 datasets. %\ph{...remarkable qualitative result...}

\begin{table}[t]
  \centering
    \caption{Clustering accuracies for several methods and datasets.  All the results of CIFAR-20 dataset were extracted from \cite{ji2019invariant}, the results of IMSAT and DEC from \cite{imsat}, the results of InfoGAN and ClusterGAN from \cite{Mukherjee2019ClusterGANL}  and the remaining from their respective papers. $^\dagger$:  average of 5 or more runs. $^\ddagger$: best of 5 runs. $\mathsection$: best of 10 or more runs. $^\|$: best of 3 runs. Results without symbols are not specified}
  \begin{adjustbox}{max width=\textwidth}
  \begin{tabular}{lccccc}
    \toprule
     & \multicolumn{5}{c}{Datasets}                   \\
    \cmidrule(l){2-6}
    Methods     & MNIST   & Onmiglot & FMNIST & CIFAR-10 & CIFAR-20 \\
    \midrule
    DEC \cite{DEC} & $84.3  ^\mathsection$ & $5.3 \pm 0.3 ^\dagger $ &  $-$  & $46.9 \pm 0.9 ^\dagger$  & $18.5$     \\
    VADE \cite{jiang2017vade}    & $94.46  ^\mathsection$  & $-$ &  $-$  & $-$   & -  \\
    InfoGAN \cite{chen2016infogan}& $89.0 ^\ddagger$  & $-$ &  $61.0 ^\ddagger$  & $-$   & -\\
    ClusterGAN \cite{Mukherjee2019ClusterGANL} & $95.0 ^\ddagger$  & $-$ &  $63.0 ^\ddagger$  & $-$   & - \\
    %CatGAN (Conv) \cite{springenberg2015catgan} & $95.63 $  & $-$ &  $-$  & $-$   & $46.9 \pm 0.9$ \\
    %DEPICT  & $96.3 $  & $-$ &  $-$  & $-$   & $-$  \\
    %JULE \cite{yang2016jule}  & $96.4$  & $-$ &  $-$  & $27.15$   & $13.7$  \\
    DAC \cite{DAC}   & $97.75^\|$  & $-$   & $-$  & $52.18^\|$  & $23.8$ \\
    %GAR \cite{kilinc2018learning}   & $98.32 \pm  0.08$ & $-$ & $-$  & $-$   & $46.9 \pm 0.9$ \\
    IMSAT (VAT) \cite{imsat}  & $98.4 \pm 0.4 ^\dagger$ & $24.0 \pm 0.9^\dagger$ & $-$ & $45.6 \pm 0.8 ^\dagger$   & - \\
    ADC \cite{Husser2017AssociativeDC} & $98.7 \pm  0.6^\dagger$ & -  & - & $ 29.3 \pm 1.5^\dagger$  & $16.0$  \\
    SCAE \cite{kosiorek2019stacked} & $98.5 \pm  0.10^\dagger$   & - & - & $33.48 \pm  0.3^\dagger$   & - \\
    IIC \cite{ji2019invariant} & $98.4 \pm  0.65 ^\dagger$   & - & - & $57.6 \pm  0.3 ^\dagger$   & $25.5 \pm 0.46 ^\dagger$\\
    \midrule
    \textbf{MPCC (Five runs)} & $98.48 \pm  0.52$   & $65.87 \pm 1.46$ & $62.56 \pm 4.16$ & $64.25 \pm  5.31$  & $ 35.21 \pm 1.69 $\\
    \textbf{MPCC (Best three runs)}  &  $98.76 \pm  0.03$   & $66.95 \pm 0.62$ &  $64.99 \pm 2.22$ & $67.73 \pm  2.50$   & $36.51 \pm 0.71$ \\
    \bottomrule
  \end{tabular}
  \end{adjustbox}

  \label{cluster_tab}
  %\vspace{-5mm}
\end{table}

\section{Discussion}

\begin{figure}[ht]%
    \begin{center}
    \subfloat[MNIST samples]{{\includegraphics[width=6.0cm]{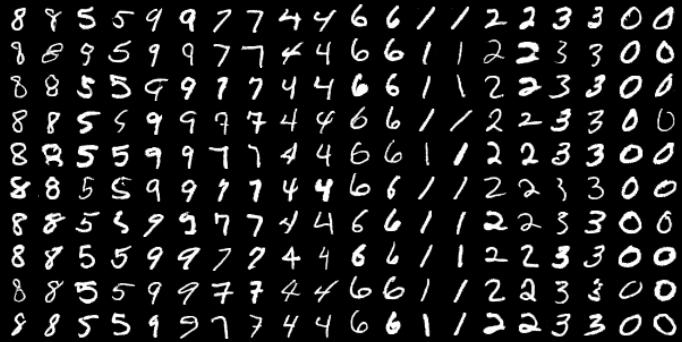} } }%
    %\qquad \qquad \qquad 
    \subfloat[CIFAR-10 samples]{{\includegraphics[width=6.0cm]{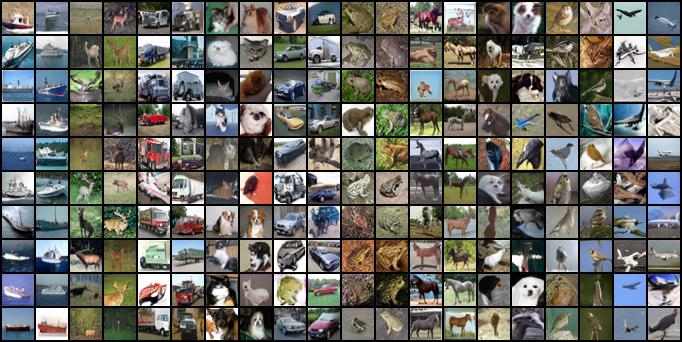} }}%
    \end{center}
    %\vspace*{-5mm}
\caption{\label{fig-uns_samples} Generated images for a) MNIST and  b) CIFAR-10 datasets, respectively.  Every two columns we set a different value for the categorical latent variable $y$.  \textit{i.e.} the samples shown correspond to a different conditional latent space $z\sim p(z|y)$.}
%\vspace{-3mm}
    \begin{center}
    \subfloat[MNIST reconstruction]{{\includegraphics[width=6.0cm]{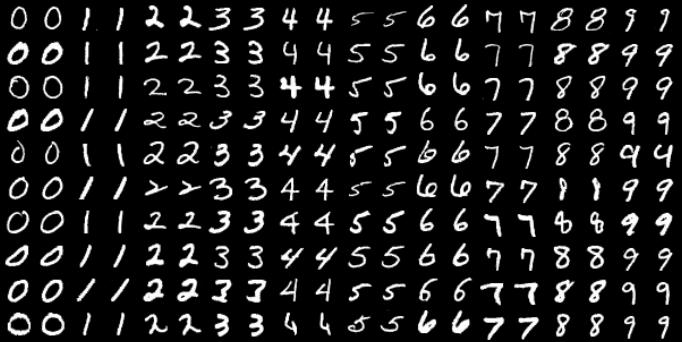} } }%
    %\qquad \qquad \qquad 
    \subfloat[CIFAR-10 reconstruction]{{\includegraphics[width=6.0cm]{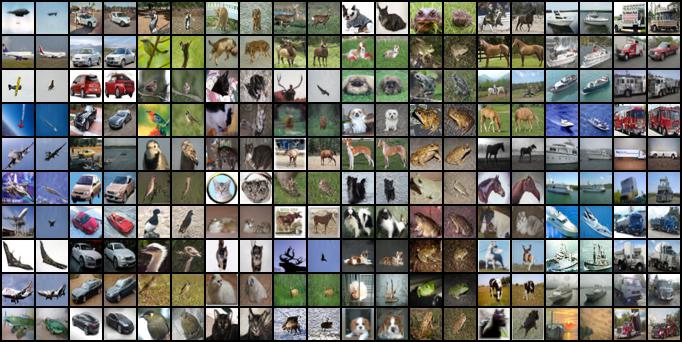} }}%
    \end{center}
    \vspace*{-5mm}
\caption{\label{fig-uns_rec} Reconstructions for a) MNIST, and b) CIFAR-10 datasets, respectively. Odd columns represent real data and even columns correspond to their reconstructions.} %The real label is used to sort the column pairs.}
%\vspace{-3mm}
\end{figure}

Our results show that MPCC 
%are backed up by its 
achieves a superior performance with respect to the SOTA on both clustering and generation quality. 
We note that the current SOTA on unsupervised and semisupervised learning relies on consistency training \cite{xie2019unsupervisedUDA} and/or data augmentation \cite{ji2019invariant}, \textit{i.e.} techniques that are complementary to MPCC and could be used to further improve our results. %that it has without using any consistency training or data augmentation. The latter techniques are the most predominant in unsupervised and semisupervised learning in terms of accuracy and are complementary with MPCC.

%The main contribution of MPCC in the context of unsupervised representation learning and clustering literature is the mathematical and optimization framework presented herein, which is backed up by its superior performance with respect to the state-of-the-art (SOTA). We note that the current SOTA on unsupervised and semisupervised learning relies on consistency training and/or data augmentation, \textit{i.e.} techniques that are complementary to MPCC and could be used to further improve our results. %that it has without using any consistency training or data augmentation. The latter techniques are the most predominant in unsupervised and semisupervised learning in terms of accuracy and are complementary with MPCC. 
To the best of our knowledge MPCC is the first deep generative clustering model capable of dealing with more complex distributions such  as CIFAR-10/20 and the first to report clustering accuracy on these datasets. Additionally we empirically prove the scalability of MPCC showing significant improvements in datasets with a larger number of classes, 20 in case of CIFAR-20 and 100 in case of Omniglot, such scalability has not been proven for the current literature on generative models \cite{jiang2017vade}, \cite{Mukherjee2019ClusterGANL}, \cite{springenberg2015catgan}, \cite{GVADE_ICCV}. %We note however that clustering accuracy varies considerably between runs and in future work we will study more robust initialization schemes and stopping criteria.

%The exceptional generation quality performance of MPCC lays in the use of parameter sharing, with an extra number of encoder updates ($E_{steps}$) and the optimization of MPCC loss function itself. These techniques are compatible with a wide range of GAN architectures and we compared them with the baseline model AIM-MPCC (NS), table \ref{comparation_tab}. This model is equivalent to a GAN without an encoder, which has no effect in the training since the prior is fixed and the encoder gradients don't flow for the generator. In terms of generation quality MPCC is robust obtaining low standard deviations results with a mean inception score and FID score of 9.49 and 6.59 respectively. \ph{I believe this paragraph is not necessary, I summarized the key aspects and added them to the next paragraph}

Our experiments show that MPCC's key innovations: GMM prior, loss function and optimization scheme (\textit{e.g.} extra encoder updates with parameter sharing) are not only relevant to achieve a good clustering accuracy but also allows us to obtain unprecedented results in terms of generation quality (Table \ref{metric_tab}). Which translates in improvements of $69.4 \%$ over the baseline (Table \ref{comparation_tab}) and $46.9\%$ over the SOTA (Table \ref{metric_tab}) in terms of FID score. We think that the exceptional generation capabilities of MPCC are related to the support that each cluster covers of the real domain. Since each cluster learns a subset of the real distribution the interpolation between two points within a cluster is smoother compared to the case where no latent separation exists. The latter is explained by the learnable shared features which exploit the similarities existing in a cluster and are not present in a fixed global prior (\textit{i.e.} ALI, AIM). %The embedding conditioned on $y$ is an explicit example of shared features that belongs to a cluster.

%We find that if we put the gaussians in a distant position from each other, \textit{i.e.} a constant hyperparameter greater than two for the prior vector $\mu_k$ the model doesn't have the capacity to learn the real distribution at least with traditional architectures. 

The high generation quality can be appreciated in Fig. \ref{fig-uns_samples} (more samples in Appendix E), where many clusters sample consistently different classes% such as boats, deer, cars, etc
. However we can still see some classes mixed in some clusters, for example in columns 7-8 with cats and dogs. MPCC also presents a competitive performance in terms of conditional distribution matching (Fig. \ref{fig-uns_rec}). The errors observed in reconstruction are semantic and similar to those observed in \cite{bigbigan}.

%In general we find good performance of MPCC across multiple datasets in particular when the model has the capacity to learn the real distribution in a complementary way with the prior parameters. Although MPCC performs worse than non-generative models in CIFAR10 and SVHN we still think that clustering algorithms based on latent data representations remain as an important advance in unsupervised learning and are particularly useful in real applications where no supervised pretrained model is available to extract useful features \cite{imsat} or there is no  clear way to perform data augmentation \cite{kilinc2018learning}.

MPCC opens the possibility of future research in many relevant topics which %we think are relevant to point out and are not explore here since they 
are out of the scope of this paper. Based on our experiments the most important extensions are: 1) Experiment with other conditional distributions $p(z|y)$, \textit{e.g.} other exponential-family distributions or other flexible distributions by bounding their entropy (Section \ref{method_section}). This can be suitable for more expressive priors as it's shown in recent work \cite{LOGAN}. 2) Experiment with imbalanced distribution of classes by changing $\phi$ accordingly, we consider this to be a relevant problem in the unsupervised setting which only a few works have addressed \cite{RDEC}. %but only a few works , to the best of our knowledge this is not a fully explored topic in the current literature. 
3) Experiment with higher resolution datasets such as ImageNet \cite{imagenet} or CelebA \cite{celeba}. Current works on clustering have not focus their attention to higher-resolution due to its complexity, MPCC is a promising approach to tackle this task from a semantic perspective \cite{bigbigan}.

%\ph{batch size mayor que 50?}

\section{Conclusions}
%\ph{llegué hasta acá}

%We consider that the main contribution of MPCC in the unsupervised representation learning and clustering literature is the mathematical and optimization framework presented herein\ph{esto es como para conclusión}. Which presents a general procedure to optimize almost any mixture of distribution, guaranteeing \ph{esto es muy fuerte} the separability of each conditional distribution. 

We developed a new clustering algorithm called MPCC, derived from a joint distribution matching perspective with a latent space modeled by a mixture distribution. %The effectiveness of MPCC is based on the mixture distribution in latent space  %The optimization is done in a decoupled way, approximating the real distribution by adversarial training and learning the posterior distributions by maximizing the likelihood \na{rev}. 
As a deep generative model this algorithm is suitable for interpretable representations, having both an inference and a generative network. The inference network allows us to infer the cluster membership and latent variables from data, while the generator performs sampling conditioned on the cluster category. 

An important contribution of this work lays in the solid mathematical and optimization framework on which MPCC is based. This framework is general and we recognize several opportunities to further enhance our model. The results obtained with MPCC improve over the state of the art in clustering methods. Most notably, our model is able to generate samples with an unprecedented high quality, surpassing the state of the art performances in both conditional and unconditional training in the CIFAR-10 dataset.

%The techniques studied $\textit{i.e.}$ mixture of distributions in the and extra encoder updates with parameter sharing are general

%The proposed model has a solid theoretical mathematical framework and our results improve over the state of the art in clustering methods. Most notably, our model is able to generate samples with an unprecedented high quality, surpassing the state of the art performances in both conditional and unconditional training in CIFAR-10 dataset. Future work includes exploring the semi-supervised setting. %and \ph{?}

\textbf{Acknowledgement} Pablo Huijse and Pablo Estevez acknowledge financial support from the Chilean National Agency for Research and Development (ANID) through grants FONDECYT 1170305 and 1171678, respectively. Also, the authors acknowledge support from the ANID’s Millennium Science Initiative through grant IC12009, awarded to the Millennium Institute of Astrophysics, MAS. We acknowledge the InnovING 2030 project of the Faculty of Engineering Sciences UACh for the support and funding provided for the development of this work. We thank Pavlos Protopapas and the IACS for hosting Nicolás Astorga through the development of this research.
\clearpage
\bibliographystyle{splncs04}
\bibliography{egbib}

\newpage
\appendix

\section{Matching marginals and conditionals is equivalent to matching joints}

In the following sections we demonstrate that for the Kullback–Leibler divergence matching the marginals and conditionals is equivalent to matching the joint distributions. In Section \ref{two_variable} we demonstrate this equivalence for the two variable case using the forward KL \cite{kingma2013vae}, this can be trivially demonstrated for the reverse KL \cite{li2019aim} interchangeably replacing the variables $z$ and $x$, and the models $p$ and $q$. In Section \ref{three_variable} we demonstrate the equivalence for the three variable case using the reverse KL (MPCC). The demonstration for the three-variable forward KL is equivalent requiring only to replace the variables $y$ and $x$, and the models $p$ and $q$.
%and in the same way that before we can directly obtain the equivalence for the forward KL interchangeably replacing the variables $y$ and $x$, and the models $p$ and $q$.

\subsection{Matching marginals and conditionals, two variable case}
\label{two_variable}

In Section 2 of the paper we show that maximizing the VAE objective (ELBO) can be interpreted as matching the conditional distributions in observed space and the marginal distributions in latent space.  Let $p(x,z)$ and $q(x,z)$ be the decoder (\textit{generation}) and encoder (\textit{inference}) joint distributions, respectively, where $x$ represents the observed variable (data) and $z$ represents the continuous latent variable. The dependencies in these distributions are expressed as  $p(x,z) =p (z) p(x|z)$ and $q(x,z) = q(z|x) q(x)$. In what follows we prove, for the two variable case, that matching conditionals and marginals is equivalent to matching the joint distributions in terms of the Kullback-Leibler divergence.
%The proposed method starts by writing the Kullback-Leibler divergence decoder (\textit{generation}) and encoder (\textit{inference}) joint distributions $p(x,z)$ and $q(x,z)$, where $x$ represents the data and $z$ represents the continuous latent variable. The dependencies in these distributions are expressed as  $p(x,z) =p (z) p(x|z)$ and $q(x,z) = q(z|x) q(x)$. 
%We can prove that the KL divergence between the joint distributions can be expressed as
Starting from the KL divergence between the joint distributions we can show that
\begin{align}
    &D_\text{KL} \left ( q(x,z) || p(x,z) \right) \nonumber \\
    =~& \int_x \int_z q(x,z) \log \frac{q(x,z)}{p(x,z)}\,dz \,dx  \nonumber \\
    =~& \int_x q(x) \int_z  q(z|x)\log \frac{q(z|x)}{p(z|x)} \,dz \,dx + \int_x q(x) \log \frac{q(x)}{p(x)} \,dx  \nonumber \\
    =~& \mathbb{E}_{x\sim q(x)} \left [D_\text{KL} \left ( q(z|x) || p(z|x) \right)  \right] + D_\text{KL} \left ( q(x) || p(x) \right),
\end{align}

\textit{i.e.} the forward KL divergence between the joints is indeed equivalent to the forward KL  between the marginals in data space plus the expected value under the data distribution of the forward KL  between the conditionals in latent space. %This is mentioned but not proven in \cite{li2019aim} (Section 4.3).

\subsection{Matching priors and conditionals for three variables \textit{a.k.a} MPCC case}
\label{three_variable}

In Section 3 of the paper we define the MPCC model starting from a joint distribution matching perspective. Let $p(x,z,y)$ and $q(x,z,y)$ be the decoder (\textit{generation}) and encoder (\textit{inference}) joint distributions, respectively, where $x$ represents the observed variable (data), $z$ represents the continuous latent variable and $y$ represents the cluster membership. 
%The proposed demonstration starts by writing the Kullback-Leibler divergence between decoder (\textit{generation}) and encoder (\textit{inference}) joint distributions $p(x,z,y)$ and $q(x,z,y)$, where $x$ represents the data, $z$ represents the continuous latent variable and $y$ represents the cluster membership. 
The dependencies in these distributions are expressed as  $p(x,z,y) = p(y) p (z|y) p(x|z,y)$ and $q(x,z,y) = q (y|z) q(z|x) q(x)$. The only assumption is that $q(y|x,z) = q(y|z)$. In what follows we prove, for the three variable case, that matching conditionals and marginals is equivalent to matching the joint distributions in terms of the Kullback-Leibler divergence.

Starting from the (reverse) KL divergence between the joint distributions we show that
\begin{align}
    &D_\text{KL} \left ( p(x,z,y) || q(x,z,y) \right) \nonumber \\
    =~& \int_x \int_y \int_z p(x,z,y) \log \frac{p(x,z,y)}{q(x,z,y)} \,dx \,dy \,dz \nonumber \\
    =~& \int_y \int_z p(z,y) \int_x  p(x|z,y)\log \frac{p(x|z,y)}{q(x|z,y)} \,dx \,dy \,dz \nonumber \\
    +~& \int_y p(y) \int_z p(z|y) \log \frac{p(z|y)}{q(z|y)} \,dz \,dy + \int_y p(y) \log \frac{p(y)}{q(y)} \,dy  \nonumber \\
    =~& \mathbb{E}_{z,y\sim p(z,y)} \left [D_\text{KL} \left ( p(x|z,y) || q(x|z,y) \right)  \right] \nonumber \\
    +~& \mathbb{E}_{y\sim p(y)} \left [D_\text{KL} \left ( p(z|y) || q(z|y) \right)  \right]  + D_\text{KL} \left ( p(y) || q(y) \right),
\end{align}
\textit{i.e.} the KL between the joints is the sum of the KL divergences for $x|(z,y)$, $z|y$ and $y$, respectively. The KL divergence is non-negative so if we match the priors and conditionals then the joints have to match too.

\section{Variational Deep Embedding (VaDE)} %revenge
\subsection{Variational Deep Embedding matches conditionals and marginals in data space} %revenge

Here we show that Variational Deep Embedding (VaDE) \cite{jiang2017vade} is in fact matching the joint distributions of the encoder and decoder by matching posteriors and marginal in the space of the observed variable $x$ (data). We start by expanding the divergence between the joint distribution of the encoder and decoder as:
\begin{align} \label{joints}
    &~D_\text{KL} \left ( q(x,z,y) || p(x,z,y) \right)\nonumber \\ 
    =&~ \int_x \int_y \int_z q(x,z,y) \log \frac{q(x,z,y)}{p(x,z,y)} \,dx \,dy \,dz \nonumber \\
    =&~ \mathbb{E}_{z,x\sim q(z,x)} \left [D_\text{KL} \left ( q(y|z,x) || p(y|z,x) \right)  \right] \nonumber \\
    +&~ \mathbb{E}_{x\sim q(x)} \left [D_\text{KL} \left ( q(z|x) || p(z|x) \right)  \right]  + D_\text{KL} \left ( q(x) || p(x) \right).
\end{align}
The first divergence in the right hand side of Eq. \eqref{joints} is
\begin{align} \label{KL1}
    D_\text{KL} \left ( q(y|z,x) || p(y|z,x) \right)  &= 
    \int q(y|x) \log \frac{q(y|x)p(z)}{p(z|y)p(y)}  \,dy \nonumber \\
    &=  \mathbb{E}_{y\sim q(y|x)} \left [ \log \frac{q(y|x)}{p(z|y)p(y)} \right] + \log p(z),
\end{align}
where we used the replacements $q(y|z,x)=q(y|x)$ and $p(y|z,x) = \frac{p(x|z)p(z|y)p(y)}{p(x|z)p(z)}$, which come from the graphical model assumptions considered in \cite{jiang2017vade}.

The second divergence in the right hand side of Eq. \eqref{joints} is
\begin{align}
    D_\text{KL} \left ( q(z|x) || p(z|x) \right)  &= 
    \int q(z|x) \log \frac{q(z|x)p(x)}{p(x|z)p(z)}  \,dz \nonumber \\
    &=  \mathbb{E}_{z\sim q(z|x)} \left [ \log \frac{q(z|x)}{p(x|z)} -\log p(z) \right] + \log p(x), \label{KL2}
\end{align}
and the third divergence in the right hand side of Eq. \eqref{joints} is
\begin{equation} \label{KL3}
    D_\text{KL} \left ( q(x) || p(x) \right)  =  \mathbb{E}_{x\sim q(x)} \left [ \log q(x) - \log p(x) \right].
\end{equation}
If we add the expectation over $q(z,x)=q(z|x)q(x)$ of Eq. \eqref{KL1} with the expectation over $q(x)$ of Eq. \eqref{KL2} and Eq. \eqref{KL3} we obtain:
\begin{equation} \label{vade-joints}
\begin{split}
&\mathbb{E}_{z,x\sim q(z,x)} \left [D_\text{KL} \left ( q(y|z,x) || p(y|z,x) \right)  \right] + \mathbb{E}_{x\sim q(x)} \left [D_\text{KL} \left ( q(z|x) || p(z|x) \right)  \right] \nonumber \\
+~ & D_\text{KL} \left ( q(x) || p(x) \right) \\
= ~ &\mathbb{E}_{q(x)} [ \log q(x) - \mathbb{E}_{z,y\sim q(z,y|x)}  [ \log p(x|z) - \log q(z|x)  - \log q(y|x) \nonumber \\ +~& \log p(z|y) + \log p(y) ]] \\
= ~ &\mathbb{E}_{q(x)} \left [ \log q(x) - \mathcal{L}_{\text{VaDE}}(x) \right],
\end{split}
\end{equation}
where $\mathcal{L}_{\text{VaDE}}(x)$ corresponds to Eq. (9) in \cite{jiang2017vade}.  This means that by maximizing VaDE's loss function one is matching the conditionals and marginals between encoder and decoder in data space. Note that the entropy of the data distribution $\mathbb{E}_{q(x)} \left [ \log q(x)\right]$ is constant during optimization. %\ph{Será necesario decir algo como esto??: The entropy term $\mathbb{E}_{q(x)} \left [ \log q(x)\right]$ is ignored during optimization}\na{yo creo que no está demás, pero yo no se si diría ignored, yo diría que es simplmente no es optimizable por que es la distribución real de los datos.} 

\subsection{Why extending VaDE to any multi-modal distribution is harder than MPCC?}

In MPCC the latent space can be naturally extended to any  mixture of distributions, the only requirement being that the entropy of each distribution component $p(z|y)$ should have a closed-form or at least a bound. In general any model decomposed by the reverse KL enjoy this property. 

Forward KL decompositions, such as the case of VAE and VaDE, need a closed-form solution for the divergence between the posterior and the prior. In VaDE this term corresponds to

\begin{equation}
    \label{divergenceVaDE}
    \mathbb{E}_{q(x)}\mathbb{E}_{q(z,y|x)}[\log q(z|x) - \log p(z|y)] = \mathbb{E}_{q(x)}\mathbb{E}_{q(y|x)}[D_{KL}(q(z|x)||p(z|y))],
\end{equation}

which has a closed-form since $q(z|x)$ and $p(z|y)$ are Gaussians. Other distributions can be used however they need to be from the exponential family and to have the same distribution \cite{exponentials}, although some exceptions exists \cite{normalgamma}, \cite{cauchy}. In addition to the exponential family requirement, a reparameterization trick is needed for the posterior distribution further limiting the distributions that can be used and requiring other forms of reparameterization \cite{imprepgrad}, \cite{genrepgrad}, \cite{beyreptrick}. 

Alternatively, adversarial training can be used to match the marginal posterior with more flexible priors. However it has been observed \cite{distributionMatching} that this kind of optimization \cite{adversarialautoencoders}, \cite{AVB} underestimates its Kullback-Leibler divergence and also worsen the likelihood of the decoder likely affecting its clustering capabilities.

\section{Details on neural network architectures}

\begin{figure}[H]%
    \begin{center}
    \subfloat[Generator block]{{\includegraphics[width=5.8cm]{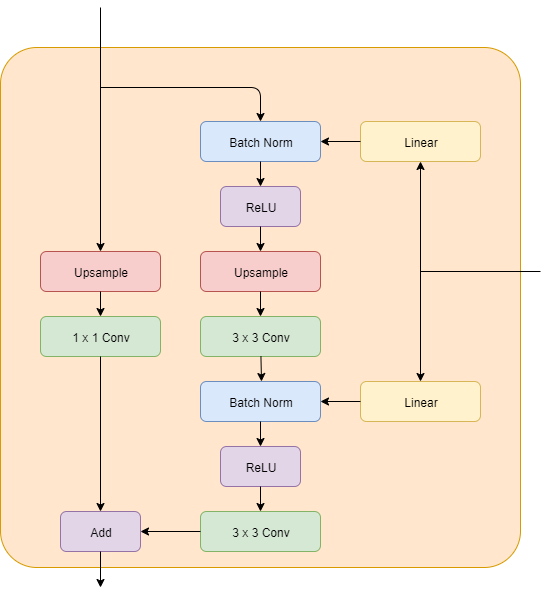} } }%
    \qquad \quad
    \subfloat[Discriminator/Encoder block]{{\includegraphics[width=4.cm]{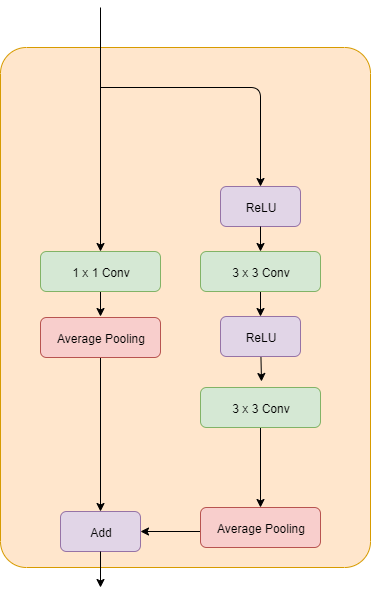} }}%
    \end{center}
    \vspace*{-2mm}
\caption{\label{blocks} Residual blocks used for MPCC generator, discriminator and encoder networks.}
\end{figure}

\begin{figure}[H]%
    \begin{center}
    \subfloat[Generator]{{\includegraphics[width=4cm]{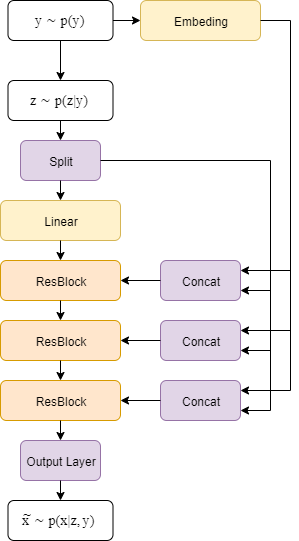} } }%
    \qquad \quad
    \subfloat[Discriminator]{{\includegraphics[width=3cm]{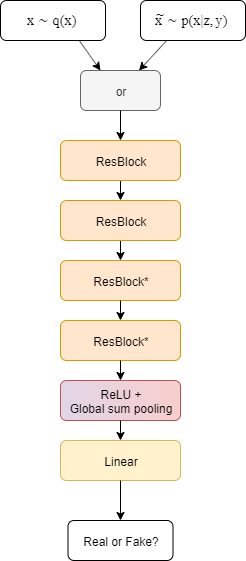} }}%
    \qquad \quad
    \subfloat[Encoder]{{\includegraphics[width=4cm]{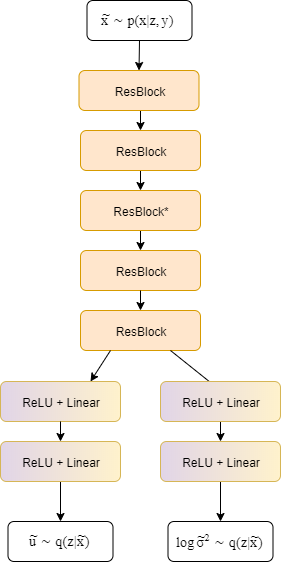} }}%
    \end{center}
    %\vspace*{-2mm}
\caption{\label{architectures} Architectures of MPCC generator, discriminator and encoder networks, respectively}
\end{figure}

In MPCC we use the BigGAN model techniques \cite{brock2018large} as a base for all our experiments. This architecture employs ResNet \cite{residual} and Spectral Normalization \cite{miyato2018spectral}. The residual block components of the generator and discriminator/encoder are shown in Fig. \ref{blocks} (a) and (b), respectively. All the $3 \times 3$ Conv use a padding equal to one while $1 \times 1$ Conv have no padding. The upsampling operation of the generator is done using bilinear interpolation. A general scheme of the generator, discriminator and encoder architecture is shown in Fig. \ref{architectures}. The first residual block of the discriminator/encoder inverts the order of the $1\times 1$ Conv and the average pooling and omits the first ReLU activation. Residual blocks with an asterisk correspond to the ones that do not perform average pooling and as a consequence they do not use $1 \times 1$ Conv. 

Fig. \ref{architectures} (a) shows the generator used for the CIFAR10/20 datasets. %which splits $z$ into chunks and after that concatenates them with the shared embedding.  %Other variants of the generator are used and the differences are observed in base of Fig. \ref{architectures} (a)  \ph{no entiendo que querias decir con "and the differences..."} \na{quería decir que bueno ocupamos tres variantes de arquitectura, pero solo pondre una para no ser tan redudante, las comparaciones se haran con la que pondre/puse}. %Es enredado hablar de eso aqui, más abajo está super claro
Fig \ref{architectures} (b) shows the unconditional architecture of the discriminator. In the case of the conditional discriminator a term $\text{Embed}(y) \cdot h $ is added where $h$ is the output of the global sum pooling (see Table \ref{discriminator_arch}). 
%Note that besides use different ways in passing the latent space, the architectures are not that different. 
We can write the architectures for all datasets in a general way as in Fig. \ref{architectures} or more specific as in Tables \ref{generator_arch}, \ref{discriminator_arch} and \ref{encoder_arch}, where $C$, $J$ and $D$ change between datasets.

As we observed in the paper (Table 3) we found an improvement in terms of sampling quality and reconstruction error when parameters between the discriminator and the encoder are shared. We experimented on the number of residual blocks shared and found that the best performance was obtained when sharing the first three residual blocks. %We note that sharing the greatest number of parameters increases performance. We shared the first three residual blocks since sharing absolutely all parameters affects negatively.

We use the Adam optimizer \cite{Adam} with its default parameters $\beta_1 = 0$ and $\beta_2 = 0.999$. We use exponential moving average (EMA) with a decay rate of $0.9999$ for the Generator for both sampling and reconstruction task. EMA is applied after the 1000th iteration. All generator, discriminator and encoder parameters use Spectral Normalization and are initialized with $\mathcal{N}(0, 0.02I)$, while an orthogonal initialization is used for prior parameters \cite{Saxe14exactsolutions}. %\cite{radford2015unsupervised}. 
With the exception of the prior parameters we use a learning rate of $2e-4$ for all networks and experiments. For evaluation we use standing statistics \cite{brock2018large} \textit{i.e.} in evaluation mode we run many times (in our case 16) the forward propagation of the generator model $\tilde{x} \sim p(x,z,y)$ storing the means and variances aggregated across all forward
passes. % storing ... lo copie directaemente de biggn u.u

%\section{Visual results on conditional sampling}

%\section{Comparison with Variational deep embedding}
%\bibliographystyle{unsrt}
%\bibliography{references}

We use three techniques depending on the dataset to deliver the latent information $z$ and $y$ into the decoder distribution $p(x|z,y)$. The first two correspond to a hierarchical latent space architecture \cite{brock2018large} which concatenate Embed($y)$ with a subset of $z$, and then a linear transformation to estimate the statistics of the batch norm layers is applied (see Fig. \ref{blocks}). The first method is the one observed in Fig. \ref{architectures} which splits the latent variable $z$ into equal chunks, delivering each one to a different part of the network. In this case we have four chunks (1 entry + 3 residual blocks). The second method is similar to first, the only difference being that all $z$ is shared and no split operation is done. The schematic of this generator is equivalent to Fig. \ref{architectures} (a) except that the purple box performs a copy instead of split operation. The third method passes all the latent $z$ as usual \cite{DCGAN} and uses conditional batch normalization \cite{condbatch_dumlolin}. This method learns embeddings conditioned on $y$ which are different for each layer, \textit{i.e.} the linear transformation in the yellow boxes of Fig. \ref{blocks} correspond to an embedding, and the shared embedding should be ignored.

\begin{itemize}
\item For CIFAR10 and CIFAR20 we use the first method since this is the default architecture used in BigGAN. For simplicity we kept this configuration for all ablation and clustering experiments with these datasets. We found that  mode collapse problems would appear if the third method is used in these datasets. The configuration of the parameters for these datasets is $C = 96$, $D = 3$ (RGB), $J = 128$ and  $\eta_p = 6 \cdot 10 ^{-4}$.

\item For datasets with simpler distributions such as MNIST and Omniglot, the third method is more stable and yields the best results. We found that if we use hierarchical latent space architectures poor results were obtained. In particular we observed that the chunks in the first method are decorrelated, which is particularly bad for simpler datasets such as MNIST and Omniglot because the network gains lot of capacity ignoring the embedding $y$ and learning the full real distribution in all the clusters. The configuration of the parameters for these datasets is $D = 1$ (grayscale), $J = 24$ and $\eta_p = 1.6 \cdot 10 ^{-3}$. For MNIST $C = 12$ and for Omniglot $C = 16$.

\item For FMNIST we observed that a poor performance was obtained with both the first and third method. %However the third method also presented lack of performance, we use the second method for this reason. 
The best results for this dataset were obtained using the second method. The configuration of the parameters for this dataset is $C = 24$, $D = 1$ (gray), $J = 16$ and $\eta_p = 1.6 \cdot 10 ^{-3}$.
\end{itemize}

We develop a Pytorch implementation for MPCC based on the implementation of BigGAN\footnote{\url{https://github.com/ajbrock/BigGAN-PyTorch}}. The IS and FID scores are calculated using the official implementations\footnote{\url{https://github.com/bioinf-jku/TTUR}}. We run each model in a GeForce RTX 2080 Ti, the amount of time that MPCC iterates depends on the dataset but it is within the range of 12-24 hours.

\newpage

\begin{table}[H]
  \begin{minipage}{.5\linewidth}
  \centering
  \begin{tabular}{c}
    \toprule
    \midrule
    $y_i \in \{ 0, \dots, K-1 \} \sim Cat(\phi)$ \\
    $z_i \in \mathbb{R}^{J} \sim  \mathcal{N}(\mu_{y_i}, \sigma^2 _{y_i} )$ \\
    Share Embed($y$) $\in \mathbb{R}^{J}$ \\
    \midrule
    Linear($J$) $\rightarrow 4 \times 4 \times 4C$    \\ \midrule
    Resblock up $4C \rightarrow 4C$    \\ \midrule
    Resblock up $4C \rightarrow 4C$    \\ \midrule
    Resblock up $4C \rightarrow 4C$    \\ \midrule
    Output Layer: \\
    BN, ReLU, $3 \times 3$ Conv $C \rightarrow 3$    \\ 
    Tanh \\

    \bottomrule
  \end{tabular}
    \vspace{1mm}

    \caption{Generator}
  \label{generator_arch}
  \end{minipage}
\begin{minipage}{.5\linewidth}
  \centering
  \begin{tabular}{c}
    \toprule
    \midrule
    $x \in \mathbb{R}^{32 \times 32 \times D}$ \\
    \midrule
    Resblock down $4C \rightarrow 4C$    \\ \midrule
    Resblock down $4C \rightarrow 4C$    \\ \midrule
    Resblock  $4C \rightarrow 4C$    \\ \midrule
    Resblock  $4C \rightarrow 4C$    \\ \midrule
    ReLU, Global sum pooling   \\ \midrule
     (linear $\rightarrow 1$) \\
      if conditional :$+$ Embed($y$)$\cdot h $ \\
    \bottomrule
  \end{tabular}
    \vspace{1mm}
    \caption{Discriminator}
  \label{discriminator_arch}
  \end{minipage}
  %\vspace{-5mm}
\end{table}

\begin{table}[H]
  \centering
  \begin{tabular}{c}
    \toprule
    \midrule
    $x \in \mathbb{R}^{32 \times 32 \times D}$ \\
    \midrule
    Resblock down $4C \rightarrow 4C$    \\ \midrule
    Resblock down $4C \rightarrow 4C$    \\ \midrule
    Resblock  $4C \rightarrow 4C$    \\ \midrule
    Resblock down $4C \rightarrow 4C$    \\ \midrule
    Resblock down $4C \rightarrow 4C$    \\
    \midrule
    Flatten \\\midrule
    $\times 2:$ Linear $(32//2^{4} \times 4C ) \rightarrow (32//2^{4} \times 4C)//2$ \\
    \midrule
    $\times 2:$ Linear $((32//2^{4} \times 4C)//2 ) \rightarrow J$ \\
    \bottomrule
  \end{tabular}
    \vspace{1mm}
    \caption{Encoder}
  \label{encoder_arch}
  %\vspace{-5mm}
\end{table}

\begin{comment}

\begin{table}[H]
   %\renewcommand\thetable{4}
  \centering
  \begin{tabular}{c}
    \toprule
    \midrule
    $y_i \in \mathbb{R} \sim Cat(\phi)$ \\
    $z_i \in \mathbb{R}^{J} \sim  \mathcal{N}(\mu_{y_i}, \sigma^2 _{y_i} )$ \\
    Embed($y$) $\in \mathbb{R}^{J}$ \\
    \midrule
    Linear($J$) $\rightarrow 4 \times 4 \times 4ch$    \\ \midrule
    Resblock up $4ch \rightarrow 4ch$    \\ \midrule
    Resblock up $4ch \rightarrow 4ch$    \\ \midrule
    Resblock up $4ch \rightarrow 4ch$    \\ \midrule
    Output Layer: \\
    BN, ReLU, $3 \times 3$ Conv $ch \rightarrow 3$    \\ 
    Tanh \\

    \bottomrule
  \end{tabular}
    \vspace{1mm}

    \caption{Generator}
  \label{metric_tab}
  %\vspace{-5mm}
\end{table}

\begin{table}[H]
   %\renewcommand\thetable{4}
  \centering
  \begin{tabular}{c}
    \toprule
    \midrule
    $x \in \mathbb{R}^{32 \times 32 \times D}$ \\
    \midrule
    Resblock down $4ch \rightarrow 4ch$    \\ \midrule
    Resblock down $4ch \rightarrow 4ch$    \\ \midrule
    Resblock  $4ch \rightarrow 4ch$    \\ \midrule
    Resblock  $4ch \rightarrow 4ch$    \\ \midrule
    ReLU, Global sum pooling   \\ \midrule
     (linear $\rightarrow 1$) \\
      if conditional :$+$ Embed($y$)$\cdot h $ \\
    \bottomrule
  \end{tabular}
    \vspace{1mm}
    \caption{Discriminator}
  \label{metric_tab}
  %\vspace{-5mm}
\end{table}

\end{comment}

\section{Optimization problems}
We observed two types of errors which restrict the architecture and the optimization techniques. Theses difficulties are particularly relevant for the CIFAR10 and CIFAR20 datasets which present the more complex distributions. We used the default parameters of the CIFAR10 architecture unless otherwise stated.

\begin{figure}[H]%
    \begin{center}
    \subfloat[Saturation problems]{{\includegraphics[width=5cm]{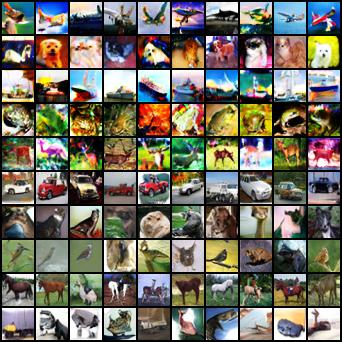} } }%
    \qquad \quad
    \subfloat[Mode collapse problems]{{\includegraphics[width=5cm]{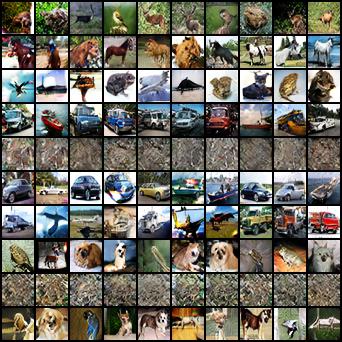} }}%
    \end{center}
    \vspace{-2mm}
\caption{\label{problems} Generated images with bad optimization setting at iteration 50000. Sub-figure (a) shows images associated with saturation problems and (b) with mode collapse problems.  Each row represents a different cluster.}
\end{figure}

The first problem is associated with the batch size. We found that we can't optimize MPCC with a big batch size while using a large learning rate of the prior parameter $\eta_p$. Note that the latter is necessary to obtain good accuracy performance as it was shown in the paper (Table 2 in the paper). The batch size is relevant to increase the IS and FID scores \cite{brock2018large}. Artifacts or saturation problems would appear when doing a small modification in the optimization. The examples shown in Fig. \ref{problems} (a)  use a batch size slightly larger than the one used in the paper (50). We observe that using a slightly larger batch size (64) with a prior learning rate of $\eta_p = 8 \cdot 10^{4}$ the results change drastically and the generated images show notable saturation.

Mode collapse is an important topic in GANs research and is the second problem that we observed in MPCC. Usually it is associated with the limitations in generation quality caused by the model, which memorize only a small part of the real distribution affecting the performance of the GAN. In MPCC the mode collapse problem can make an entire cluster collapse. Setting $D_{step} = 4$ solves this problem partially for a large amount of models and is sufficient to obtain good performance. In Fig. \ref{problems} (b) we show samples from a model trained with $bs = 64$ and $\eta_p = 2\cdot 10 ^{-4}$ where we can see how a mode collapse problem looks in MPCC. We observed that when using a large prior learning rate $\eta_p = 6 \cdot 10 ^{-4}$ this problem would regularly appear after 150,000 iterations. %This problem would occur regularly when the model iterates for more than 150000 iterations largely enough when using a relatively large prior learning rate $\eta_p = 6 \cdot 10 ^{-4}$.%, where we can observe this problem in the iteration range of $150000 - 200000$. 
This doesn't occur for the best configuration of MPCC (the one reported in the paper) and setting $\eta_p = 2 \cdot 10^{-4}$ even after a large number of iterations, however we would like to increase $\eta_p$ further as we observed that it correlates with better clustering accuracy (Table 2 in the paper). %in our work a large prior learning rate is relevant for good performance.

\newpage 

\section{Additional qualitative results}

\begin{figure}[H]
  \vspace{5mm}

\begin{center}
    \includegraphics[width=0.8\textwidth]{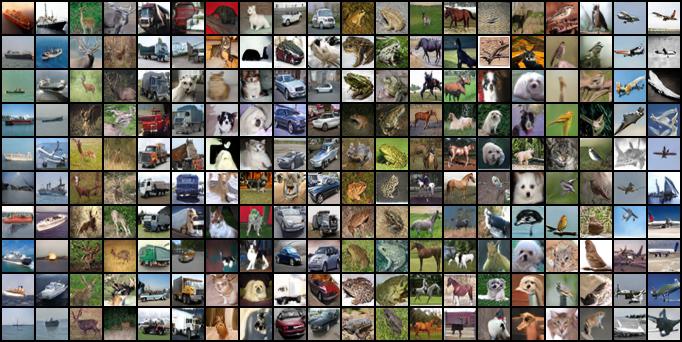}
\end{center}
 \caption{\label{Add_samples_cifar} Generated images for the CIFAR-10 dataset.  Every two columns we set a different value for the categorical latent variable $y$.  \textit{i.e.} the samples shown correspond to a different conditional latent space $z\sim p(z|y)$.}
 %\vspace{-5mm}
\end{figure}

\begin{figure}[H]
\begin{center}
    \includegraphics[width=0.8\textwidth]{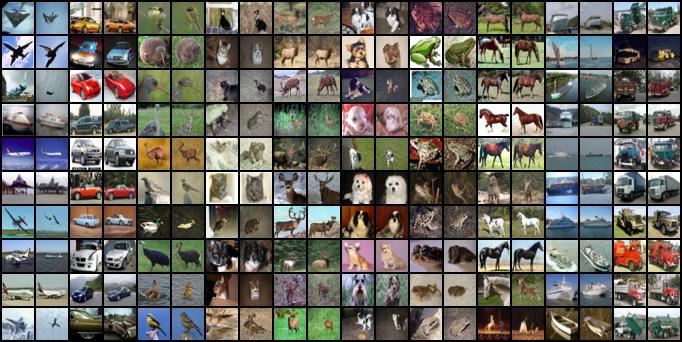}
\end{center}
  %\vspace{-5mm}
 \caption{\label{Add_reconstruction_cifar} Reconstructions for rhe CIFAR-10 dataset. Odd columns represent real data and even columns correspond to their reconstructions. The real label is used to sort the column pairs.}
 %\vspace{-5mm}
\end{figure}

\begin{figure}[H]
 \vspace{20mm}

\begin{center}
    \includegraphics[width=0.8\textwidth]{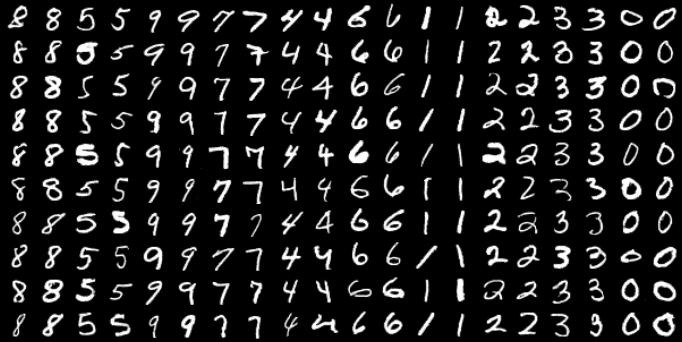}
\end{center}
  %\vspace{-5mm}
 \caption{\label{Add_samples_mnist} Generated images for the MNIST dataset.  Every two columns we set a different value for the categorical latent variable $y$.  \textit{i.e.} the samples shown correspond to a different conditional latent space $z\sim p(z|y)$.}
 %\vspace{-5mm}
\end{figure}

\begin{figure}[H]
 \vspace{10mm}

\begin{center}
    \includegraphics[width=0.8\textwidth]{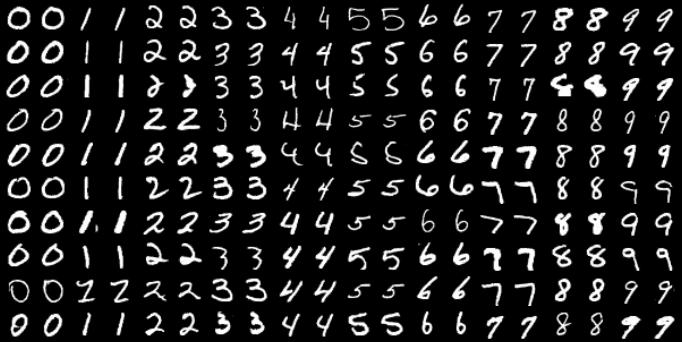}
\end{center}
  %\vspace{-5mm}
 \caption{\label{Add_reconstruction_mnist} Reconstructions for the MNIST dataset. Odd columns represent real data and even columns correspond to their reconstructions. The real label is used to sort the column pairs.}
 %\vspace{-5mm}
  \vspace{10mm}

\end{figure}

\begin{figure}[H]
  \vspace{40mm}

\begin{center}
    \includegraphics[width=0.8\textwidth]{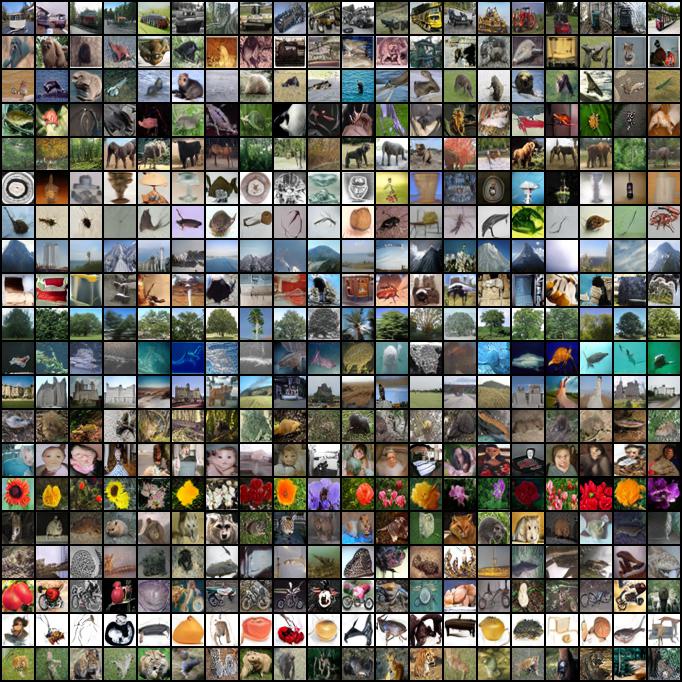}
\end{center}
  %\vspace{-5mm}
 \caption{\label{Add_samples_C20} Generated images for CIFAR-20 dataset.  In every row we set a different value for the categorical latent variable $y$, \textit{i.e.} the samples shown correspond to a different conditional latent space $z\sim p(z|y)$.}
  \vspace{30mm}

 %\vspace{-5mm}
\end{figure}

\begin{figure}[H]
\begin{center}
    \includegraphics[width=0.8\textwidth]{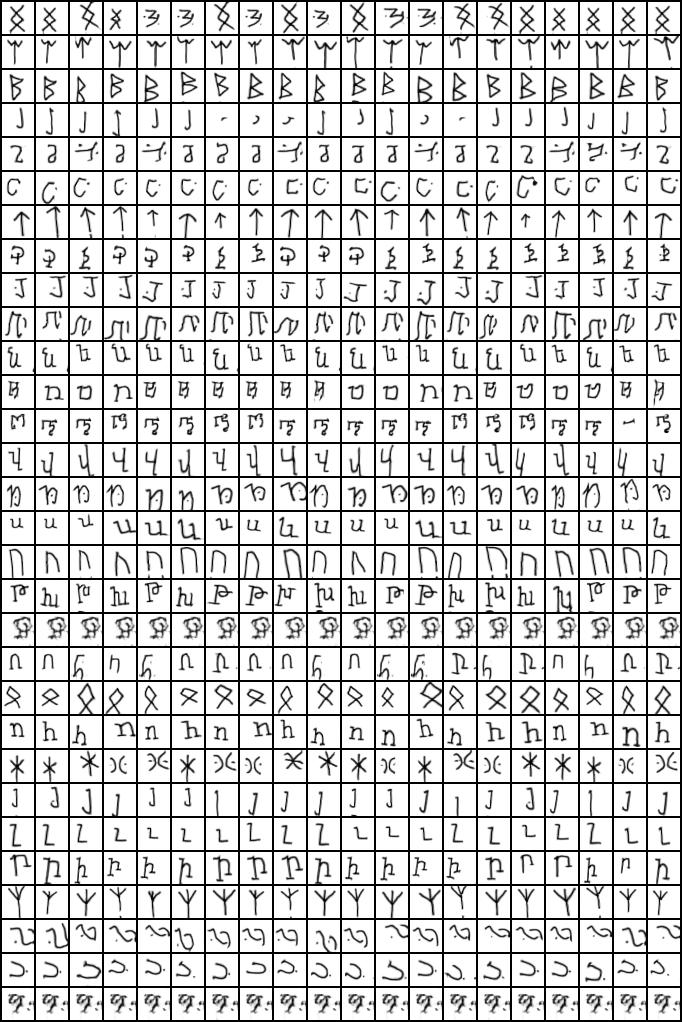}
\end{center}
  %\vspace{-5mm}
 \caption{\label{Add_samplesOmni} Generated images for Omniglot dataset.  In every row we set a different value for the categorical latent variable $y$, \textit{i.e.} the samples shown correspond to a different conditional latent space $z\sim p(z|y)$. 30 cluster were randomly chosen.}
 %\vspace{-5mm}

\end{figure}

In this section we provide additional reconstructions and samples for the CIFAR-10 dataset in Figures \ref{Add_samples_cifar} and \ref{Add_reconstruction_cifar}, and for the MNIST dataset in \ref{Add_samples_mnist} and \ref{Add_reconstruction_mnist}. To give more insight about MPCC's capacity we also include samples for datasets with a high number of classes, CIFAR-20 and Omniglot in Figures \ref{Add_samples_C20} and \ref{Add_samplesOmni} respectively.

\end{document}